\documentclass[10pt,twocolumn,letterpaper]{article}

\usepackage[pagenumbers]{cvpr}      %

\usepackage{array}   %

\definecolor{cvprblue}{rgb}{0.21,0.49,0.74}
\usepackage[pagebackref,breaklinks,colorlinks,allcolors=cvprblue]{hyperref}

\title{Persona Lenses: What Makes a Vision--Language Model See Through Its Persona's Eyes?}
\hypersetup{pdftitle={Persona Lenses: What Makes a Vision-Language Model See Through Its Persona's Eyes?}, pdfauthor={Yoonho Shin}}

\author{Yoonho Shin\\
LG Uplus
}

\begin{document}
\widowpenalty=10000
\displaywidowpenalty=10000
\clubpenalty=10000
\makeatletter\@clubpenalty=10000\makeatother
\maketitle

\begin{abstract}

Vision--language models have been used to simulate where a user looks on a web
page. However, a persona in the prompt reaches the simulated gaze only partly.
We measure this reach as the difference in one model's gaze between two
MatrAIx personas that differ in a single attribute. We introduce persona
lenses, which place the persona between the model and the page: a text lens (a
two-step prompt in which the model first writes what the persona would look
at), a pixel lens (marks on the region matching the persona's interests) and a
weight lens (a dwell reward gated to keep the scanpath natural). Using
Qwen3-VL (4B and 32B) and Qwen3.5-9B on generated and live pages, we compare
the lenses with one another and with scale, the 32B model with no lens. It has
been confirmed that the text lens separates the personas strongly but is not robust
to profile changes that should not move gaze. The
pixel lens separates the personas less but is unaffected by attributes
unrelated to gaze. At 4B and 9B, the weight lens separates the personas on interests
held out from training, with no extra step on a new page. The 32B model separates the personas with the
base prompt alone. Therefore, the choice of a lens depends on cost and model
size.

\end{abstract}

\begin{figure*}[t]
  \centering
  \includegraphics[width=\linewidth]{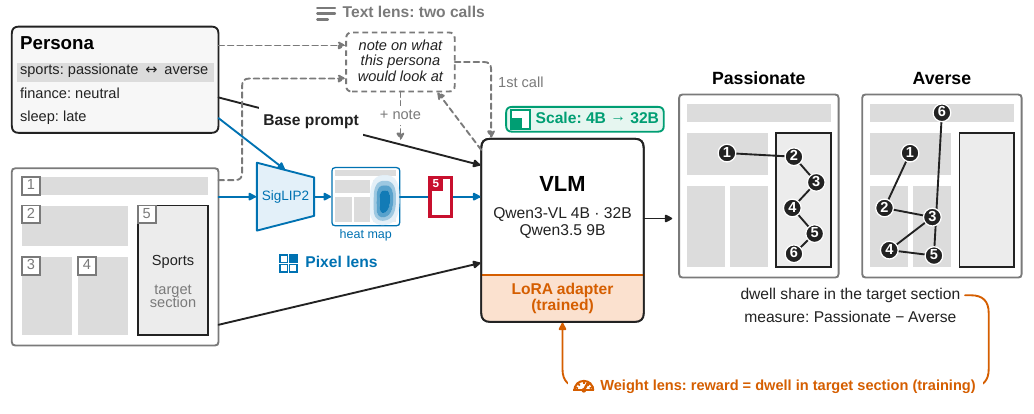}
  \caption{Overview. A vision--language model receives a page screenshot and one of two persona versions that differ in one interest, and returns a simulated scanpath. The measure is the target-section dwell share of the Passionate version minus that of the Averse version. Three persona lenses and model scale change this gaze. The text lens (dashed) calls the same model twice on the screenshot, first for a note on what this persona would look at, then for the scanpath with the note added to the base prompt. The pixel lens encodes the page and each interest with SigLIP2, finds the interest's region as a heat-map blob and draws the marker of a liked region in red. The weight lens is a reward on target dwell. The lenses act only on the scanpath prediction, and the note is always written without the pixel lens.}
  \label{fig:routes}
\end{figure*}

\section{Introduction}
\label{sec:intro}

Where users look on a web page tells a service which content is seen and which is passed over, so models that predict this attention are used to judge a layout before it is changed~\cite{ueyes2023chi,uniar2024,gdd_attention2024}. Users with different interests look at different content~\cite{semanticsalience2019,interestgaze2023,webviewing2004}, so this picture is needed for each group of users. An eye-tracking study gives it~\cite{ueyes2023chi},
but such a study cannot be run for every page and every group of users. Scanpath models fitted to eye-tracking data take observed gaze or a
demographic group as input~\cite{eyeformer2024,premap2025}. Persona
agents~\cite{uxagent2025} read a persona written in text but act on the HTML and do not predict where the user looks. To our knowledge, no study has tested whether a persona written in text
reaches the gaze. Given a
screenshot with Set-of-Mark tags~\cite{som2023} and a persona profile, a
vision--language model (VLM) returns a scanpath. However, a persona in the prompt reaches the simulated gaze only
partly, as in other persona-prompting
studies~\cite{helpfulassistant2023,personaeffect2024,personasel2026,agentpersonabench2026}.
No public gaze dataset labels viewers with personas.
Therefore, this paper asks how far a persona reaches the gaze of one model.

Each persona has two versions, Passionate and Averse, that differ in one interest, and the persona effect is the gap in their share of dwell time on that interest's section. A small VLM is cheap enough to run on many pages and personas, but with the base prompt its persona effect at 4B is less than a third of that at 32B (\cref{sec:scale}). A two-step prompt, in which the model first writes what the persona would look at, brings the 4B model near the 32B model with the base prompt. However, it needs an extra generation per page and separates personas that differ only in an attribute unrelated to gaze, such as knitting (\cref{sec:diagnosis}). It is also unclear whether it changes the attention or only the scanpath.

We call any method that places the persona between the model and the page a persona lens (\cref{fig:routes}). We treat the two-step prompt as a text lens, propose a pixel lens and a weight lens, and compare the three with model scale, which needs none. The pixel lens leaves the model unchanged and highlights the tags of the region that an image encoder matches to the persona's interests. Because it reads only the interests, an attribute unrelated to gaze cannot change its marks. The weight lens uses reinforcement learning~\cite{visualrft2025} to reward persona-aligned dwell under a naturalness gate, so that a new page needs no extra step.
\Cref{sec:method} describes the lenses, \Cref{sec:experiments} compares them with scale, and \Cref{sec:mechanism} analyzes what each lens changes inside the model.

\section{Related Work}
\label{sec:related}

\paragraph{Visual prompting.}
Visual prompting steers a VLM with marks on the image~\cite{vpsurvey2024}. Numbered tags~\cite{som2023}, red
circles~\cite{redcircle2023}, blur outside a mask~\cite{fgvp2023} and
query-guided attention maps~\cite{attnprompt2024} need no training, while free-form drawings need fine-tuning~\cite{vipllava2023}.
Attention prompting and the red circle are closest to our pixel lens. However, they serve tasks with one right answer in which the task chooses the region, and the mark's appearance changes the results~\cite{vpfragile2025}. In our pixel lens, the persona chooses the region, so one page needs different marks for different readers.

\paragraph{Gaze and scanpath prediction.}
Models fitted to eye-tracking data~\cite{ueyes2023chi} predict
scanpaths~\cite{eyeformer2024,deepgaze35vl2026}, attention maps on
designs~\cite{uniar2024,gdd_attention2024}, attention under a prompt or
task~\cite{attendanything2026,tdsal2026} and search for a named
target~\cite{seekui2026}. Concurrent work trained partly on UEyes predicts one saliency map
per image for all viewers~\cite{openvam2026}. Human gaze varies with the viewer~\cite{semanticsalience2019,webviewing2004,agegaze2018} and follows interest, preference and goals~\cite{interestgaze2023,gazepreference2003,goalattention2007}. Personalized models condition on a person's own fixations~\cite{personalsal2024,beyondaverage2024,fixate2026,fewshotscan2025} or on an age and gender group~\cite{premap2025}. None of these models takes a persona written in text, and fitting one needs gaze data for each persona. Instead, we steer a VLM that reads the persona text.

\paragraph{Persona conditioning.}
Persona corpora provide profiles for simulated users, and we draw ours from one~\cite{matraix2026}. Persona agents interact with interfaces~\cite{uxagent2025,amuser2026} or evaluate them~\cite{perceptui2026,simab2026}, but none of them predicts where the user looks. Moreover, a persona in the text prompt changes a model's behavior only partly. Its effect is
often weak, with no gain in factual accuracy~\cite{helpfulassistant2023} or
survey agreement~\cite{personasurvey2026} and a small shift in simulated
judgments~\cite{personaeffect2024}. A persona can also cause unintended changes. Setting
one attribute shifts unset ones~\cite{personasel2026}, traits partly drop
across environments~\cite{agentpersonabench2026}, and simulated populations
collapse~\cite{distfirst2026} or turn into caricatures~\cite{caricature2026}.
Our gaze results show both weak and unintended effects (\cref{sec:diagnosis,sec:lens,sec:internalise}).

\paragraph{Reinforcement learning and self-distillation.}
Visual reinforcement fine-tuning rewards a correct region or
click~\cite{visualrft2025,uir1_2025,guig1_2025}. Spatial attention is trained inside the reinforcement
loop~\cite{gazevlm2026} or toward
objects the model names~\cite{selfsal2026}, and robot policies learn from
human gaze~\cite{gazeprompts2026}. Reinforcement learning also helps a model act on its knowledge~\cite{greedyagents2025} and, after supervised fine-tuning, on its persona~\cite{customerr12025}, but no reward targets where a persona looks.
Another route into the weights is on-policy self-distillation~\cite{distillcontext2022,system2distill2024,guiopsd2026}, in which a student matches a teacher that sees a privileged input, such as a marked screenshot, answer crops, a drawn box, coordinates or a text hint~\cite{guisd2026,trustteacher2026,thinkwoimages2026,ioupd2026,whereopd2026,ptdpo2026}. This input is the same for every user, and only a text assistant's fixed identity contains a persona~\cite{edgeopd2026}. Our pixel lens is a privileged input that changes with the persona, and we use it at test time instead of distilling it.

\section{Method: Persona Lenses and Model Scale}
\label{sec:method}

With the base text prompt alone, a persona reaches the simulated gaze only
partly: the 4B model barely separates two opposite personas on live pages,
and in every model the gaze moves less than when the model first writes what
the persona would look at (\cref{sec:diagnosis,sec:scale}). This two-step prompt, the text lens, reaches
further, but its note must be written per page with the screen in view, and it
also moves the gaze for attributes that have no counterpart on the page
(\cref{sec:diagnosis}). Therefore, we propose two more persona lenses that make the gaze follow the persona's interests: the pixel lens marks the screenshot, and the weight lens trains the weights. We compare the three lenses and model scale with one another, with the text lens as the comparison point.

\subsection{Task and Text Lens}
\label{sec:baselines}

\begin{figure}[t]
\noindent\fbox{\parbox{\dimexpr\linewidth-2\fboxsep-2\fboxrule\relax}{\raggedright\scriptsize\ttfamily
\textbf{Base prompt, system turn.} You are simulating the visual attention of
a specific person browsing a web page. The person: \{persona\}. [\ldots]
Simulate, as this specific person, the order in which their eyes would land on
the marked elements, and how long they would dwell on each. [\ldots] Respond
ONLY with a JSON object:
\{"scanpath": [\{"id": <marker>, "dwell\_ms": <int>\}, ...], [\ldots]\}.\\[2pt]
\textbf{Base prompt, user turn.} TASK for this person: \{task\}. Simulate this
person's scanpath over the marked screenshot now.\\[3pt]
\textbf{Text lens adds a first turn.} This is a web page with numbered markers
on its elements. The person: \{persona\}. In one or two sentences, describe how
this person's characteristics and interests will shape which kinds of elements
they look at on this page and for how long. Answer in plain prose, without
marker numbers.\\[2pt]
\textbf{Then, prepended to the user turn.} Note about this person, from your
own assessment of this page: \{note\}}}
\caption{The prompts. The base prompt is the system and user turns alone;
\{persona\} is one sentence of attribute phrases, such as ``aged 18-24;
\ldots; Passionate about Sports; \ldots'', and a pair
flips one interest phrase. The text lens adds a first turn, with the plain-marker
screenshot, that writes the note and prepends it to the user turn; the pixel and
weight lenses keep the base prompt.}
\label{fig:prompt-box}
\end{figure}

\paragraph{Task.}
The model receives a page screenshot with Set-of-Mark tags and a persona profile, and answers with an ordered list of fixations, which we call the simulated gaze (\cref{fig:prompt-box}). Each fixation is a marker number with a dwell time of 80 to 3000\,ms, a range that the prompt sets and that holds 97.6\% of human fixations in UEyes~\cite{ueyes2023chi}. Two versions of a persona, the Passionate and Averse poles, differ in one interest, and a lens or scale is judged by how far apart their simulated gazes lie on the target section (\cref{sec:setup}).

\paragraph{Base prompt and text lens.}
All conditions share the screenshot and the marker set. The
\textbf{base prompt} (\emph{off}) gives the persona with plain markers. The
\textbf{text lens} is the comparison point. The model first sees the
screenshot and the persona and writes a note on what this
person will look at on this page, without being told the interest under test.
The note is then given back to the model at the start of the gaze prompt (\cref{fig:prompt-box}). A note written once per persona
without the screen keeps only 16--50\% of the text-lens effect, so
we write the note per page with the screen in view, and in a turn of its own (Appendix~A).

\subsection{Pixel Lens}
\label{sec:method-lens}

The \textbf{pixel lens} acts in the pixels and leaves the weights and the prompt untouched: it marks the screenshot from the interests in the profile alone, without being told which one is under test, so the two poles' screenshots differ only through the flipped interest. It finds the region of each interest with the SigLIP2 image--text encoder~\cite{siglip2}, as the heat-map blobs that best match the interest text. Then it draws the markers of liked regions in red with a 3-px box and the markers of disliked regions in translucent gray.

\subsection{Weight Lens}
\label{sec:method-weights}

\begin{figure}[t]
\centering
\includegraphics[width=\linewidth]{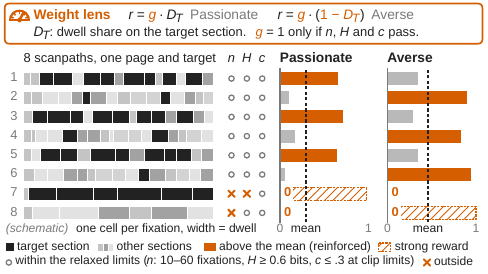}
\caption{Gated dwell reward (schematic). The same eight scanpaths are scored under both poles for illustration, but in training each pole's prompt gets its own eight. A scanpath that passes the gate earns $D_T$ under Passionate and $1-D_T$ under Averse, and none here is reinforced (orange) under both. Scanpaths 7 and 8 fail the gate and get 0, while the strong reward (hatched) would rank each first under one pole.}
\label{fig:weightlens}
\end{figure}

Since the text and pixel lenses act on every page at test time, the \textbf{weight lens} trains the weights so that the gaze follows the persona with no extra step on a new page (\cref{fig:weightlens}).

\paragraph{Reward.}
For a scanpath $s$, $D_T(s)$ is the share of its dwell on the target section $T$. The reward is
\begin{gather}
r(s)=g(s)\cdot\begin{cases}D_T(s) & \text{Passionate pole,}\\ 1-D_T(s) & \text{Averse pole,}\end{cases}
\label{eq:reward}\\
g(s)=\mathbf{1}[\,s\text{ is valid},\ \underline{n}\le n\le\overline{n},\ H\ge\underline{H},\ c\le\overline{c}\,],
\label{eq:gate}\\
\textstyle H=-\sum_k p_k\log_2 p_k,
\label{eq:entropy}\\
\textstyle c=\frac{1}{n}\sum_i\mathbf{1}[d_i\le 80\text{ or }d_i\ge 3000],
\label{eq:clip}
\end{gather}
where $\mathbf{1}[\cdot]$ is 1 if the condition in it holds and 0 otherwise, so $g(s)=1$ only when all four conditions hold, under- and overbars mark lower and upper limits, $s$ is valid if it is well formed and not repetitive, $n$ is the fixation count, $p_k$ is the dwell share of section $k$ or of the rest of the page, and $d_i$ is the dwell time of fixation $i$ in ms. Since the two poles reward opposite behavior on the same page and target, only reading the persona raises the gated reward. The \emph{strong reward} uses $(\underline{n},\overline{n},\underline{H},\overline{c})=(5,40,0,1)$.

\paragraph{Training.}
We train rank-16 LoRA~\cite{lora2021} adapters on the 4B and 9B models for 500 steps with group relative policy optimization (GRPO)~\cite{deepseekmath2024}. The model samples eight scanpaths per prompt at temperature 1, the advantage of each is its reward standardized within the group, and a Kullback--Leibler term keeps the policy near the untrained model. Training uses the base prompt, 30 personas, 9 pages outside the evaluation set, 12 interests and the text-size need (rewarded on the smallest elements), with each interest only on the pages that contain its section. No reward targets fashion or travel, which we hold out to test whether the adapter reads the persona or recalls the rewarded interests (\cref{sec:internalise}), and no training profile mentions them, although six training pages show a fashion section and one a travel section.

\paragraph{Naturalness gate.}
The strong reward collapses the scanpath into an all-or-nothing response to the target (\cref{sec:internalise}). Therefore, a naturalness gate keeps the validity test but replaces the 5\nobreakdash--40 fixation range with limits on $n$, the dwell spread $H$ and the clip share $c$ in \cref{eq:gate}. Our first gate takes $(\underline{n},\overline{n},\underline{H},\overline{c})=(15,60,1.0,.2)$ from the untrained 4B model's outputs on an earlier version of the evaluation pages, not from human gaze, and there is no validation split. The untrained model passes it in 76\% of its outputs, but training the 4B model under it had no effect on the held-out interests. As our protocol allowed for that case, we relaxed the limits once to $(10,60,0.6,.3)$, and registered them before the new runs (Appendix~D). We use this \emph{relaxed gate} at 4B and 9B.

\subsection{Model Scale}
\label{sec:method-scale}
Scale is not a lens. It changes neither the screenshot nor the weights but gives the base prompt to the 32B model (\cref{sec:setup}) and asks whether size alone suffices.

\section{Experiments}
\label{sec:experiments}

\begin{figure*}[t]
  \centering
  \includegraphics[width=\linewidth]{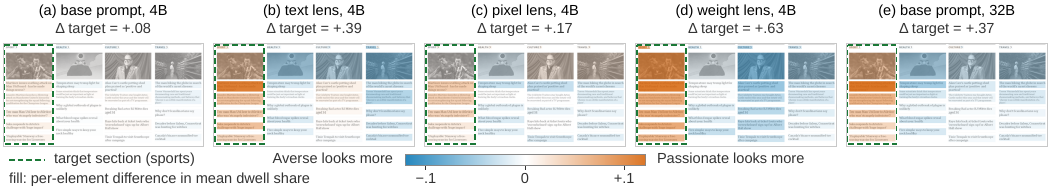}
  \caption{The lenses on one live page (BBC, sports), each alone on the base prompt. The weight lens is a relaxed-gate adapter whose reward covers sports and whose training uses these 30 personas, so panel (d) is not held out (\cref{tab:internalise}). Color: Passionate minus Averse dwell share per element (30 personas, 2 seeds); $\Delta$: target-section pole difference.}
  \label{fig:diffheat}
\end{figure*}

We measure the lenses and scale under one protocol, and \cref{fig:diffheat} shows each of them on one live page. Full prompts, training settings, detailed results and further analyses are in the supplementary material.

\subsection{Experimental Setup}
\label{sec:setup}

\paragraph{Stimuli.}
All stimuli are $1280\times800$ screenshots with at most 60 Set-of-Mark tags.
Generated pages give every interest a labeled section with an exact box, so
the controlled comparisons use them, and live pages test whether the results
hold on real sites. The generated-page set has eight pages per interest, all
generated except a live Naver page for fashion and finance, which the
live-page set also holds.

\paragraph{Live-page evaluation set.}
The set has 87 cells for 6 interests on 61 screenshots of 19 sites, where a
\emph{cell} is one (page, interest) pair. Its \emph{earlier set} (44 cells) comes from an earlier round of this study, and its \emph{later set} (43 cells) mostly from other sites, with shopping sites and more technology, cooking and travel cells. The target boxes were drawn under a
registered rule, and those of the later set were committed before any model output was seen.

\paragraph{Personas.}
We draw 30 synthetic people from the MatrAIx persona
corpus~\cite{matraix2026}, each described in a name-free sentence by 22 or 23
attributes.

\paragraph{Measures.}
Target-section dwell share is the fraction of dwell inside the section of the
flipped interest. A fixation belongs to the section whose
box contains the center of its element, and the denominator is total dwell,
including dwell outside every labeled section. The pole difference is the Passionate minus
the Averse share, paired on (page, persona, seed). On the generated set, the test for each model and interest pools 480 pairs (8 pages $\times$ 30 personas $\times$ 2 seeds), so its $p$-value treats the pages as fixed. A live cell has 60 pairs. We test
each condition against the base prompt unless noted, with a one-sided paired Wilcoxon test~\cite{wilcoxon1945} and, where stated, a Benjamini--Hochberg correction~\cite{benjamini1995} whose
adjusted $p$ we write as $q$. We also count the
pages with a positive mean increment and test the 30
persona means.

\paragraph{Preregistration.}
Hypotheses, measures, sample sizes and thresholds were committed before data collection to protocol files, which will be released with the code (Appendix~A), and no registered threshold was changed except the one gate relaxation that the protocol allowed (\cref{sec:method-weights}). Every
measure, criterion or analysis added after data collection is marked as such
in the supplement. We report failed predictions, including a rejected rule that predicts which live cells fail, and an amended check (Appendices~G and~K).

\paragraph{Models.}
To test the lenses on a varied lineup, we run Qwen3-VL-4B-Instruct and Qwen3-VL-32B-Instruct (4B and 32B)~\cite{qwen3vl2025}, Qwen3.5-9B (9B)~\cite{qwen35_9b} of a newer, natively multimodal generation, and InternVL3.5-8B~\cite{internvl35_2025}, whose language model starts from Qwen3-8B. \Cref{tab:internalise} adds two closed models run through an API, GPT-6.1-terra and GPT-6-luna (Appendix~K).
Unless noted, each input is sampled with two seeds.

\subsection{What a Prompt Reaches}
\label{sec:diagnosis}

We first ask how far a persona in the prompt reaches the simulated gaze. For interests, the base prompt reaches it only partly: the pole difference in target dwell share on sports is $.076$ at 4B, $.191$ at 9B and $.296$ at 32B (\cref{sec:scale}). The text lens reaches much further, by $.11$ to $.29$ on the generated pages (\cref{tab:lens-scale}). However, its note has to be written per page with the screen in view (\cref{sec:baselines}), and it also moves the gaze for gaze-irrelevant attributes, as shown below. Therefore, we keep the text lens as the comparison point for the other lenses and scale.  

\paragraph{Irrelevant attributes and human-likeness.}
The text lens is not robust to profile changes that should not move gaze. We flipped gaze-irrelevant attributes: sleep schedule, rock music, knitting,
AI for documents and the partly related Korean food.
The text lens differs from the base prompt in 12 of 15 tests (10 of 12 without Korean food), by 5 to 12\% of its sports effect. Merely swapping two profile phrases moves its gaze about as much, and far more than the base prompt's (post hoc, Appendix~J). It also lowers human-likeness: measured against UEyes viewers~\cite{ueyes2023chi}, it changes ScanMatch~\cite{scanmatch2010} and saliency CC by $-.018$ to $-.045$ at 4B and 9B. All eight confidence intervals reach below the registered bound of $-.02$, and three include zero.

\subsection{Pixel Lens: Effect of Encoder-Guided Marks}
\label{sec:lens}

\begin{table}[t]
\centering\footnotesize
\setlength{\tabcolsep}{2pt}
\caption{Increments over the base prompt on generated pages. Each value is the pole difference in target-section dwell share under a condition minus that under the base prompt, except Pixel$-$rand.\ (pixel lens minus random marker). True sect.\ is the true-section marker, and recovery (rec.)
is $(\text{pixel}-\text{base}) \div (\text{true}-\text{base})$. The
text lens column comes from a separate run with its own base prompt. Every entry
has one-sided paired Wilcoxon $p<.05$ over 475--480 pairs except the two marked ns.}
\label{tab:lens-scale}
\begin{tabular}{@{}llcccc@{}}
\toprule
Interest & Model & True sect. & Pixel lens (rec.) & Pixel$-$rand. & Text lens \\
\midrule
Sports
 & 4B  & $+.069$ & $+.013$ (.18) & $+.009^{\mathrm{ns}}$ & $+.288$ \\
 & 9B  & $+.048$ & $+.032$ (.66) & $+.051$ & $+.128$ \\
 & 32B & $+.053$ & $+.010^{\mathrm{ns}}$ (.19) & $+.019$ & $+.146$ \\
Fashion
 & 4B  & $+.066$ & $+.028$ (.43) & $+.028$ & $+.185$ \\
 & 9B  & $+.070$ & $+.031$ (.44) & $+.034$ & $+.111$ \\
 & 32B & $+.075$ & $+.016$ (.21) & $+.033$ & $+.117$ \\
Finance
 & 4B  & $+.075$ & $+.066$ (.87) & $+.066$ & $+.218$ \\
 & 9B  & $+.088$ & $+.062$ (.71) & $+.079$ & $+.152$ \\
 & 32B & $+.068$ & $+.026$ (.38) & $+.060$ & $+.190$ \\
\bottomrule
\end{tabular}
\end{table}

\paragraph{Effect of the pixel lens.}
Two controls share the pixel lens's rendering: the \textbf{true-section marker} applies it to the true section boxes, and the \textbf{random marker} emphasizes or dims as many randomly chosen elements as the pixel lens, with the same signs.
\Cref{tab:lens-scale} shows that the pixel lens significantly raises the pole difference
above the base prompt in 8 of 9 entries and above the random marker in 8 of 9.
Of the 34 significant entries, 32 remain so in a test over the 30 persona means and 25 in a test over the 8 page means (Appendix~C). At the page level, 2 of the 8 significant Pixel lens entries remain so against 6 of the 8 significant Pixel$-$rand.\ entries, with similar page-cluster bootstrap intervals.
The random marker alone stays below the base prompt in 7 of 9 entries and at most $+.004$ above it in the other two, never significantly, so emphasis on random elements does not raise the pole difference. The recovery is $.43$ to $.87$ at 4B and 9B except on 4B sports ($.18$), but at most $.38$ at 32B, where the true-section marker is still significant.

\paragraph{Generalization and limits.}
The pixel lens carries over to another model family, another encoder and further interests, except where the encoder misses the region. On InternVL3.5-8B it is significant in all three interests and
recovers $.73$, $.59$ and $.93$, and with CLIP ViT-L/14-336~\cite{clip2021} in place of SigLIP2,
its recovery is at least $.5$ at 4B and 9B except on sports at 4B (Appendix~C). On four further interests it is significant in 11 of 12 entries, but on technology it recovers only $.09$, $.31$ and $.02$ of the true-section marker at 4B, 9B and 32B, which is effective there, so the weakness lies in finding the region. 

\paragraph{Comparison with the text lens.}
\Cref{tab:lens-scale} shows that the text lens separates personas several times more than the pixel lens in every generated-page entry. On InternVL3.5-8B the
two are of the same order on fashion and finance. Unlike the text lens (\cref{sec:diagnosis}), the pixel lens is unaffected by the gaze-irrelevant attributes: it differs from the base prompt in 1 of 15 tests (knitting at 9B).

\paragraph{Pixel lens on top of text prompts.}
On generated pages the pixel lens adds on top of text prompts at 4B and 9B, and less often at 32B. On top of the text lens it adds in 3 of 3 interests at 4B and 9B and in at most 1 of 3 at 32B in both runs, and on InternVL in 3 of 3.
On a text
ceiling, an author-written sentence naming the target interest, it adds in 6 of 6 tests at 4B and 9B (3 of 6 with
pages as units) and in 1 of 3 at 32B. A three-step prompt that
also plans the gaze beats the text lens in 9 of 9 tests, and the pixel lens
adds to it in 7 of 9.

\paragraph{Marker-naming control.}
Dwell alone cannot rank the lenses: the gaze output is a list of marker numbers, so a sentence naming them
nearly dictates the list. A registered control gives the pixel lens's selection as text instead of pixels. On a plain screenshot, one sentence names the markers it would highlight and dim. As text, the selection adds $.301$--$.494$ at 4B and $.135$--$.224$ at 9B, above the pixel lens in every entry. At 4B the control changes the target attention defined in \cref{sec:mechanism} by only $+.003$ and $+.021$ on fashion and travel, and $.99$ of its effect remains under the late-layer mask. At 4B the sentence also lowers the median fixation count from 30 to 15, while the pixel lens keeps 27. At 9B the count does not fall (27 to 28). Naming the dimmed markers tells the model to avoid them, but the pixel lens's gray markers do not. Therefore,
we also measure attention, irrelevant attributes and scanpath
shape. We read the larger effect of text, even over the true-section marker, the same way: a note states the preference, while a marker's color must be interpreted.

\begin{table}[t]
\centering\footnotesize
\setlength{\tabcolsep}{2.6pt}
\caption{Live-page cells (\% of the set) where the pole difference exceeds that under the base prompt
(one-sided paired Wilcoxon $p<.05$). The indented row also beats the random marker.}
\label{tab:realpages}
\begin{tabular}{@{}lccc@{\hspace{11pt}}ccc@{}}
\toprule
 & \multicolumn{3}{c}{Both sets (87 cells)} & \multicolumn{3}{c}{Later set (43 cells)} \\
\cmidrule(lr){2-4}\cmidrule(lr){5-7}
Condition & 4B & 9B & 32B & 4B & 9B & 32B \\
\midrule
Text lens                  & 70\,{\scriptsize(80)} & 76\,{\scriptsize(87)} & 65\,{\scriptsize(75)} & 35\,{\scriptsize(81)} & 38\,{\scriptsize(88)} & 34\,{\scriptsize(79)} \\
Pixel lens                 & 34\,{\scriptsize(39)} & 30\,{\scriptsize(34)} & 28\,{\scriptsize(32)} & 21\,{\scriptsize(49)} & 17\,{\scriptsize(40)} & 13\,{\scriptsize(30)} \\
\quad and above random     & 26\,{\scriptsize(30)} & 21\,{\scriptsize(24)} & 27\,{\scriptsize(31)} & 14\,{\scriptsize(33)} & 12\,{\scriptsize(28)} & 13\,{\scriptsize(30)} \\
True-section marker        & 57\,{\scriptsize(66)} & 65\,{\scriptsize(75)} & 67\,{\scriptsize(77)} & 31\,{\scriptsize(72)} & 35\,{\scriptsize(81)} & 32\,{\scriptsize(74)} \\
Text + pixel lens          & 74\,{\scriptsize(85)} & 80\,{\scriptsize(92)} & 67\,{\scriptsize(77)} & 37\,{\scriptsize(86)} & 39\,{\scriptsize(91)} & 30\,{\scriptsize(70)} \\
\bottomrule
\end{tabular}
\end{table}

\paragraph{Live pages.}
The same order holds on the live-page set. \Cref{tab:realpages} shows that the text lens succeeds in 75--87\% of the 87 cells and the pixel lens in 32--39\%,
with similar shares on the later set alone (79--88\% and 30--49\%), and that the two together
gain at most 4 cells over the text lens alone. The
random marker alone beats the base prompt in 8, 8 and 4 cells, near the 4.4
expected by chance. On live pages the pixel lens is limited by finding the target region: the true-section marker, which differs only in the highlighted region, succeeds in 66--77\% of the
cells, while the pixel lens misses technology regions and succeeds in few shopping cells. Re-scoring with target boxes redrawn by an LLM-based annotation agent changes no count by more than one cell (Appendix~G). As on generated pages, the marker-naming control beats the pixel lens (Appendix~J).

\subsection{Weight Lens: Transfer to Held-out Interests}
\label{sec:internalise}

\paragraph{Transfer and scanpath style of the strong reward.}
The weight-lens evaluation set has 10 unseen personas, 4 unseen generated and 2 live portal pages, with fashion and travel held out. Each variant has five training seeds, but the strong reward at 9B has four, and six adapters are retrained copies with the same recipe, since their checkpoints were lost (Appendix~D). \Cref{tab:internalise} shows that the strong reward transfers to the held-out interests at 4B
and 9B, as registered ($q<.001$ in every seed).
However, the scanpath collapses, more so at 9B. At 4B, the median
fixation count falls from $27$ to $7$--$13$, and 56--89\% of the Averse-pole scanpaths place
no dwell on the target, against 14\% for the base model. Therefore, these adapters give an all-or-nothing response, a caricature also diagnosed in simulated survey populations~\cite{caricature2026}.

\paragraph{Effect of the naturalness gate.}
The relaxed gate transfers in every seed at 4B and 9B ($q<.001$). At 4B it also meets the registered size and style criteria in 3 of 5 seeds, and in three its concentration lies within $.06$ of the 32B base model's (\cref{tab:internalise}). With the base prompt alone, its pole difference on the held-out interests is above that of the untrained model with the text lens in every seed at both sizes ($+.369$ and $+.264$ against $+.260$ and $+.156$ at 4B, $+.552$ and $+.589$ against $+.271$ and $+.259$ at 9B), and at 4B it is as large as that of the 32B base model ($+.366$ and $+.212$).
 At 9B its concentration is higher, so the gate needs per-model tuning.

\begin{table}[t]
\centering\footnotesize
\setlength{\tabcolsep}{2pt}
\caption{Transfer to the held-out interests on the weight-lens evaluation set (\cref{sec:internalise}), with the base prompt. Held-out $\Delta$ is the pole difference on fashion and travel, which are never rewarded and appear in no training profile ($\pm$ s.d.\ over training seeds). Conc.\ (concentration) is the dwell share in the most-visited section. $\dagger$ adds the text lens at test time in a separate run with its own base prompt. The last two rows are closed models run through an API, at the server's default temperature and with no seed.}
\label{tab:internalise}
\begin{tabular}{@{}lccc@{}}
\toprule
 & \multicolumn{2}{c}{Held-out $\Delta$} & \\
\cmidrule(lr){2-3}
Variant & Fash. & Trav. & Conc. \\
\midrule
Base 4B            & $+.066$ & $+.030$ & $.37$ \\
\quad + text lens$^\dagger$ & $+.260$ & $+.156$ & $.46$ \\
Strong reward      & $+.604\pm.120$ & $+.383\pm.125$ & $.56$--$.84$ \\
Relaxed gate       & $+.369\pm.092$ & $+.264\pm.070$ & $.51$--$.63$ \\
\midrule
Base 9B            & $+.182$ & $+.169$ & $.43$ \\
\quad + text lens$^\dagger$ & $+.271$ & $+.259$ & $.50$ \\
Strong reward 9B   & $+.735\pm.149$ & $+.711\pm.019$ & $.91$--$.97$ \\
Relaxed gate 9B    & $+.552\pm.067$ & $+.589\pm.036$ & $.67$--$.75$ \\
\midrule
Base 32B           & $+.366$ & $+.212$ & $.49$ \\
\midrule
Base GPT-6.1-terra & $+.555$ & $+.420$ & $.55$ \\
Base GPT-6-luna    & $+.266$ & $+.179$ & $.50$ \\
\bottomrule
\end{tabular}
\end{table}

\paragraph{Costs of the adapters.}
The adapters have two costs. First, like the text lens (\cref{sec:diagnosis}), they respond to attributes that should not move gaze. In every seed, the relaxed-gate 9B adapter separates personas that differ only in rock music, knitting or the partly related Korean food ($q<.05$), at 41 to 76\% of the sports effect. The 4B adapters do so less often, on Korean food in 3 and 4 of 5 seeds (22\%) and on rock music and knitting in at most 3. Second, the pixel lens is weaker on two of the three adapters. Averaged over seeds, it is significant in 1.6, 3.4 and 1.6 of 6 interests on the 4B strong-reward, 4B relaxed-gate and 9B relaxed-gate adapters, against 3 and 4 on the base 4B and 9B, and on the 9B adapter it lowers the sports effect in 4 of 5 seeds (Appendix~D).

\begin{table}[t]
\centering\footnotesize
\setlength{\tabcolsep}{4pt}
\caption{Models fine-tuned with supervision on UEyes viewers and relaxed-gate 4B scanpaths, which answer with coordinates. Trav.\ gives the gain in persona effect on travel over the same-size model trained on UEyes alone ($q<.05$ in every seed). Img.-spec.\ is the image-specific share as a ratio to that of humans, and its values differ by less than their s.d.\ over seeds ($.09$). Values are means over three training seeds, one at 32B (Appendix~I).}
\label{tab:human}
\begin{tabular}{@{}lccc@{}}
\toprule
Model & Trav. & ScanMatch & Img.-spec. \\
\midrule
Human viewers & --- & $.427$ & $1.00$ \\
Joint model, 4B & $+.201$ & $.433$ & $1.00$ \\
Joint model, 9B & $+.219$ & $.430$ & $.93$ \\
Joint model, 32B & $+.263$ & $.425$ & $.92$ \\
\bottomrule
\end{tabular}
\end{table}

\paragraph{Human gaze in the weight lens.}
Human gaze can be added to the weight lens by supervised fine-tuning, without a reward (\cref{tab:human}): adapters at 4B, 9B and 32B trained on
UEyes~\cite{ueyes2023chi} viewers and relaxed-gate 4B scanpaths as targets match each image's
viewers about as specifically as humans do, while their persona effect on both held-out interests exceeds that of the
same-size model trained on human scanpaths alone ($q<.05$ in every seed; fashion in Appendix~I).

\subsection{Scale: Effect of Model Size}
\label{sec:scale}

\begin{figure}[t]
  \centering
  \includegraphics[width=\linewidth]{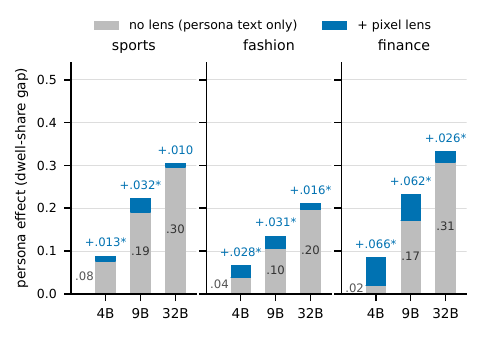}
  \caption{Scale on generated pages. Each panel is one interest, and the bars give the pole difference (persona effect) at 4B, 9B and 32B without a lens (gray) and the increment the pixel lens adds (blue, starred at one-sided paired $p<.05$). The increment is significant in 8 of 9 entries and does not grow from 4B to 32B.}
  \label{fig:scaling}
\end{figure}

\paragraph{Effect of model size on the pole difference.}
With the base prompt alone, a larger model separates the personas more. On the generated-page set the pole difference rises from
$.076$, $.039$ and $.019$ at 4B to $.296$, $.196$ and $.308$ at 32B on sports,
fashion and finance (\cref{fig:scaling}). \Cref{tab:internalise} shows the same rise on the
weight-lens evaluation set: on the held-out interests the 32B value is 5.5 and 7.1 times the 4B one ($q<.001$), and two closed models of unknown size also separate the personas on this set. On its two live portal pages the 32B value is $+.225$ against $+.019$ at 4B and $+.098$ at 9B (80 pairs, Appendix~E). However, post hoc, about 70\% of the 32B effect on the generated pages is lost on copies without topic names (Appendix~C).

\paragraph{Text and pixel lenses at 32B.}
At 32B the text and pixel lenses add less than at 4B. In a separate run on the generated-page set,
the text lens adds $.120$--$.158$ at 32B against $.166$--$.280$ at 4B over the three interests.
The text ceiling of \cref{sec:lens} adds more than the text lens at 32B but less at 4B (the lens reaches $.61$--$.66$ and $1.26$--$1.59$ times the ceiling), so the 32B model follows the persona text well. However, the cut of the note at 600 characters shortens 72\% of the 32B notes against 1\% at 4B (post hoc, Appendix~A). At 32B the pixel lens adds less than a tenth of the base effect
and little on top of text or on live pages (\cref{sec:lens}).

\paragraph{Scanpath style of the larger model.}
The 32B model reaches its persona effect without the collapse of the strong reward (\cref{sec:internalise}): its scanpath has 25 fixations against 27 at 4B and
a concentration of $.49$ against $.37$.

\section{Analysis: Attention to the Target}
\label{sec:mechanism}

Dwell is the model's output, and text alone can set it (\cref{sec:lens}). To find whether the personas also differ in where the model looks, we read its attention and intervene on it.

\paragraph{Measurement.}
For each condition and seed we regenerate the scanpath (temperature $.7$).
A teacher-forced pass over it reads the share of the generated tokens'
attention to image tokens that falls on the target section, in excess of the section's area share. We call the last quarter of the layers the late layers, and
the late-layer pole difference of this excess the target attention.
To intervene, we mask the attention from the generated tokens to the target section in one quarter of the layers or in all (the full mask) and renormalize over the remaining keys. Every masked output stays valid.

\begin{figure}[t]
\centering
\includegraphics[width=\linewidth]{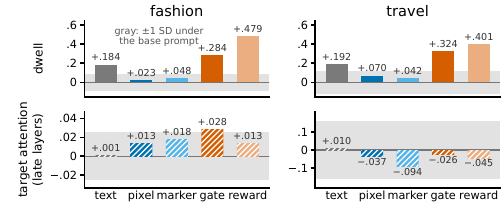}
\caption{Change over the base prompt at 4B on five pages of the weight-lens evaluation set (two fashion, three travel), in the pole difference of target dwell share (top) and in the target attention (bottom). Gray bands: $\pm 1$ standard deviation across pairs under the base prompt (post hoc, Appendix~J). Text, pixel and marker: text lens, pixel lens and true-section marker (five generation seeds). Gate and reward: relaxed-gate and strong-reward adapters (three each). One page gives most travel attention values.}
\label{fig:attn-dwell}
\end{figure}

\paragraph{Text lens.}
At 4B the text lens changes the dwell that the model reports on the target, not how much its late layers attend to it.
On the five pages of \cref{fig:attn-dwell}, it moves dwell by $+.184$ and $+.192$ on fashion and travel but the target attention by $+.001$ and $+.010$. In units of each quantity's standard deviation under the base prompt, these are $2.2$ and $1.6$ against less than $.1$.
Under the late-layer mask it keeps $.86$ and $.46$ of its dwell increment on fashion and travel (mean $.66$, above the registered $.5$).
A registered re-measurement on the generated-page set finds a rise in 5 of the 9 entries of \cref{tab:lens-scale}, with a maximum of $+.011$.

\paragraph{Pixel lens.}
At 4B the pixel lens and the true-section marker do not act mainly through late-layer attention to the target either.
On the five pages they raise the target attention on fashion in all five seeds ($+.013$ and $+.018$), by more standard deviations than they raise dwell, but on travel one page makes it fall in all five ($-.037$ and $-.094$).
Post hoc, in the first two quarters of the layers both raise this pole difference on both interests in all five seeds, by 3 to 13 standard deviations, against at most $.6$ in any quarter for the text lens.
Under the late-layer mask the true-section marker keeps $.96$ of its travel dwell increment, and the pixel lens less than $.5$ in 2 of 5 seeds, fewer than the registered four.
In the re-measurement the pixel lens raises the target attention in 5 entries, and its mean falls in all five seeds in the other 4.
The true-section marker raises it in 4 and lowers it in 2.

\paragraph{Weight lens and scale.}
At 4B and 32B the weight lens and scale raise dwell on the held-out interests more than the target attention, which rises on both interests only at 9B (\cref{fig:attn-dwell}, Appendix~F). In the same units the two 4B adapters raise dwell by $2.8$ to $5.7$ and the target attention by at most $1.1$, on fashion only.
The target attention of the 32B base model is $+.026$ and $+.009$, against $+.010$ and $+.055$ at 4B. At 9B the
strong reward raises it by $+.141$ and $+.110$ over the base model and the
relaxed gate by about half that ($+.074$ and
$+.051$), with one adapter each. 

\paragraph{Intervening on attention.}
To test causation, we intervene as in prior work~\cite{gazeheads2026,crossmodalflow2025,causalflow2026,lingrecode2026}. At 4B the weight lens relies on direct attention to the target, though on no single quarter of the layers, while scale keeps over half of its effect without it. Under the late-layer mask the two 4B adapters (seed 0) keep $.65$ and $.70$ of their dwell difference, above the registered $.5$. Under the full mask they keep $.36$ and $.07$, and the 32B model $.62$ and $.55$.
Six of twelve registered predictions on attention failed, mostly on the size or the layers of the change (Appendix~F).

\section{Conclusion}
\label{sec:conclusion}

A persona in the prompt moves the simulated gaze of a VLM only partly. We compared persona lenses in text, pixels and weights, and model
scale, with the text lens as a strong baseline, under preregistered protocols
across four models and generated and live pages. The text lens
separates personas without training but is not robust to profile changes that should not move gaze. The pixel lens is unaffected by attributes unrelated to gaze, but its effect is small, less than a tenth of the base effect at 32B, and limited by region finding,
which future work could train. The relaxed-gate weight lens
transfers to held-out interests at 4B and 9B ($q<.001$ in every seed), needs no extra
step on a new page and reaches a 32B-like concentration at 4B in 3 of 5 seeds. Model scale is not a lens. At 32B the base prompt alone separates the personas, and the lenses add less. At 4B no lens acts mainly through late-layer attention to the target. In practice: use the text lens to avoid training, add the pixel lens at 4B and 9B where the
encoder finds the target, and train the weight lens to avoid a per-page step.

\paragraph{Limitations.}
We do not test where a person with either persona looks: public gaze
datasets do not label viewers with personas, so human-likeness is checked only against
unlabeled viewers~\cite{ueyes2023chi}, the untrained 4B model
stays below randomly paired humans on ScanMatch~\cite{scanmatch2010},
and the text lens lowers it further
(\cref{sec:diagnosis}). The text and weight lenses separate personas on gaze-irrelevant attributes
(\cref{sec:diagnosis,sec:internalise}),
and the pixel lens misses technology regions and succeeds in few
shopping cells (\cref{sec:lens}). Marker numbers named in a prompt nearly
dictate the dwell share, a limit of the Set-of-Mark format. The
mask removes only direct attention (\cref{sec:mechanism}). The
personas are synthetic profiles~\cite{matraix2026}, the pages are static
single-viewport screenshots, the live set is small, the weight lens and
scale are tested only on Qwen models, and training is not bitwise
reproducible. Persona lenses could help pages capture one group's attention.

\section*{LLM Usage Statement}

We used a large-language-model (LLM) coding and writing assistant (Claude, Anthropic)
under the author's direction throughout this work. The research ideas are entirely the author's own. The assistant wrote the experiment code
and ran it on remote GPUs, wrote the analysis scripts
and drafted the experiment protocols that the author approved before data
collection. It also searched and checked the related literature, and drafted and edited
the text, tables and figures. Every number in the paper was produced by these
scripts from the logged trials.

LLM-based agents also drew the target section boxes of the live pages under the
registered rule. A separate LLM-based annotation agent, which did not see these boxes, redrew them in 63 cells.

The photographs on the generated pages were produced by the text-to-image model Z-Image-Turbo~\cite{zimage2025} from one-line scene descriptions, and the assistant helped the author judge whether each photograph shows its topic (Appendix~A).

The personas come from the MatrAIx corpus~\cite{matraix2026}, whose attribute
schema was curated with LLM assistance. Our profiles are drawn from its
synthetic part, which is sampled from a dependency graph over the schema, and
we render them into sentences with a fixed template.

The outputs of the evaluated models, including the note of the text lens,
are objects of study.

{
    \small
    \bibliographystyle{ieeenat_fullname}
    \bibliography{refs}
}

\clearpage
\maketitlesupplementary
\setcounter{table}{0}\setcounter{figure}{0}
\renewcommand{\thetable}{S\arabic{table}}\renewcommand{\thefigure}{S\arabic{figure}}
\makeatletter\renewcommand{\theHtable}{supp.\arabic{table}}\renewcommand{\theHfigure}{supp.\arabic{figure}}\makeatother

\noindent This supplement gives the details that the main paper points to. For every
registered test behind a result of the main paper it gives the verdict,
including the failed predictions, and every departure from the registered
rule. Analyses added after data collection are marked as such. Three
registered rules were amended: the gate was relaxed once as the protocol
allowed (Appendix~D), the marker-reading check of the closed models was
amended after its first version failed and before any design call
(Appendix~K), and the persona-effect criterion of the human-gaze model was
replaced before that model was scored (Appendix~I), where it fails in both
forms.

The protocol files and the code will be released. They also register tests that the main paper does not use.
Appendices~B and~H report some of these, and the others are not reported
here. Tables, figures and sections cited without an S are those of the main
paper. The generated-page set is the page set of Table~1, and the weight-lens
evaluation set is that of Sec.~4.4. \cref{tab:guide} lists what each appendix gives and where
the main paper uses it.

\begin{table}[h]
\centering\footnotesize
\setlength{\tabcolsep}{2.5pt}
\caption{Guide to the supplement. The appendices keep the letters by which the main paper cites them and do not follow the order of its sections.}
\label{tab:guide}
\begin{tabular}{@{}l>{\raggedright\arraybackslash}p{0.585\linewidth}l@{}}
\toprule
App. & Contents & Main \\
\midrule
A & System prompt, task sentences, answer length, text lens, generated pages, live-page cells & Secs.~3.1, 4.1, 4.3, 4.5 \\
B & Perceptual and cognitive traits, declarative probe, input format & --- \\
C & Pixel lens, blob detector, page-level tests, pages without topic names, other model family and encoder, further interests, earlier lens & Secs.~3.2, 4.3, 4.5 \\
D & Retrained adapters, training settings, gate and its relaxation, text lens in training, irrelevant attributes & Secs.~3.3, 4.4 \\
E & Registered scale run, live portal pages, text size & Sec.~4.5 \\
F & Attention measure, target attention by model, masks, 12 registered predictions & Sec.~5 \\
G & Cell sets, technology and shopping cells, random marker, re-annotation, earlier round & Secs.~4.1, 4.3 \\
H & Personas with two interests & --- \\
I & Joint training with human gaze & Sec.~4.4 \\
J & Three-step prompt, marker naming, attention, irrelevant attributes, floor check, UEyes & Secs.~4.2, 4.3, 5 \\
K & Closed models: access, settings, input checks, refusals, results & Secs.~4.1, 4.4, 4.5 \\
\bottomrule
\end{tabular}
\end{table}

\appendix
\section{Setup Details}
\label{supp:setup}

\paragraph{Prompt.}
\begin{figure}[t]
\noindent\fbox{\parbox{\dimexpr\linewidth-2\fboxsep-2\fboxrule\relax}{\raggedright\footnotesize\ttfamily
You are simulating the visual attention of a specific person browsing a web
page.\\[3pt]
The person: aged 18-24; based in Sub-Saharan Africa; Woman; with a Some
college education; works in Agriculture; Comfortable with technology; in the
Student life stage; their visual vs verbal thinking is Mixed; their attention
span is Average; their reading vs watching is Prefers video; their vision is
Low vision; their high-contrast need is Prefers; their large-text need is No;
their assistive technology is Magnifier; their cognitive load capacity is
Moderate; their device context is Low-bandwidth; their accessibility needs is
Visual; their daily screen time is 2-4 hrs; \underline{Passionate about
Sports}; Interested about Fashion; Neutral about Personal finance.\\[3pt]
You will see a screenshot in which interactive and content elements are marked
with small numbered tags. Simulate, as this specific person, the order in which
their eyes would land on the marked elements, and how long they would dwell on
each. Stay faithful to what THIS person (given their characteristics) would
attend to, not what an average user would do. Judge from what you SEE in the
screenshot.\\[3pt]
Respond ONLY with a JSON object:\\
\{"scanpath": [\{"id": <marker number>,\\
\hspace*{2em}"dwell\_ms": <int>\}, ...],\\
\hspace*{1em}"stop\_reason": "<satisfied|bored|\\
\hspace*{2em}found\_target|would\_scroll>"\}\\[3pt]
Rules: 5 to 25 fixations; use only marker numbers visible in the screenshot;
the same number may repeat; dwell\_ms between 80 and 3000.}}
\caption{System prompt. The profile is persona 0 of the 30 at the Passionate
pole of the sports pair, rendered with the phrase of each attribute. The underlined phrase is the one the pair flips, and
the Averse twin reads ``Averse about Sports'' with every other phrase
unchanged. This persona's attention-condition attribute is None, so its phrase
is omitted and the sentence has 21 phrases.}
\label{fig:prompt}
\end{figure}
The system prompt is given in \cref{fig:prompt} with one rendered profile in
the place of the persona.
The user turn attaches the marked screenshot and reads ``TASK for this
person: \{task\}. Simulate this person's scanpath over the marked screenshot
now.''

A page without a task of its own receives a default task sentence, which
differs between scripts and page types. The weight-lens scripts (training, the
weight-lens evaluation set and the attention analysis on its pages) use
``Browse this page for a few minutes and see what catches your
eye.'' All other runs use ``Catch up on today's news for a few minutes.'',
including all 43 later live-page cells with the 8 cells on Amazon and Walmart
pages. However, in these runs the 4 generated shop pages and 17 live portal
screenshots, which hold 26 of the 44 earlier cells, receive the browse
sentence instead, with ``the store'' or ``this portal page'' in place of
``this page''.

\paragraph{Length of the answers.}
We keep every scanpath as the model wrote it, even beyond the upper bound of
25 fixations, which 41\% to 64\% of the base-model scanpaths exceed on the
weight-lens evaluation set. An answer that reaches the cap of 1000
new tokens ends inside the JSON object, and the parser keeps its complete
fixations. The raw answers were not logged. Therefore, after data collection we
estimated from the parsed scanpaths that 10\% to 28\% of the 840 scanpaths of
the 4B base model, 0.1\% to 12\% of the 9B and 1\% to 4\% of the 32B reached
the cap. A trial whose answer does not parse or names no valid marker has no
dwell share, and the (page, persona, seed) pair that loses a pole drops out of
the test. This is why some entries of Table~1 of the main paper have fewer
than 480 pairs.

\paragraph{Text lens (two-step prompt).}
The first step uses the note prompt of Figure~2 of the main paper, in a single
user turn with no system prompt and with the plain-marker screenshot.
Except for the closed models of Appendix~K, the note is decoded greedily
(temperature 0, seed 0) with at most 150 new tokens and is then cut at 600
characters. In a count made after data collection, 1\%, 12\% and 72\% of the
4B, 9B and 32B notes on the generated-page set of Table~1 of the main paper
are exactly 600 characters long, and 2\%, 24\% and 87\% of the notes on the 87
live-page cells. Nearly all of these notes stop inside a sentence, so the cut
shortened them. The note is written once for each page, persona and pole, and
within a run it is reused unchanged in every condition and in both seeds.
Therefore, the pixel lens enters only in the second step, whose screenshot
carries it.

\paragraph{Checks of the note.}
We ran two checks of the note. First, a note written once per persona without the
screenshot keeps at most half of the effect. Reused on every page, it keeps $.27$, $.23$ and $.33$ of
the effect of the screen-seen note at 4B (sports, fashion, finance), $.29$,
$.37$ and $.50$ at 9B and $.16$, $.38$ and $.50$ at 32B. Only 2 of 9 entries
reach $.5$.
Second, we added 78 and then 278 attributes from the MatrAIx
schema~\cite{matraix2026} to every profile. At 278 attributes the
screen-seen note keeps
$.74$ to $1.35$ of its effect over the nine entries, and every entry stays
significant. The pixel lens keeps $.71$ to $.83$ at 4B, $.77$ to $1.68$ at 9B
and at least $1.13$ at 32B. The screen-free note weakens in
every entry, and the pole difference of the base prompt itself shrinks as the
profile grows
(4B sports from $.069$ to $.034$, 32B sports from $.306$ to $.167$).

\paragraph{Generated pages.}
The 12 generated pages (sp-001 to sp-012) are 4 portals, 4 shops and 4 news
front pages from our page generator. Each has three to six sections
under literal topic headings (Sports, Fashion, Finance, Food, Tech, Health or
Travel), and every image slot holds a photograph that
Z-Image-Turbo~\cite{zimage2025} generated
from a one-line scene description written for its card.

A photograph in which an OCR check finds text is drawn again with another
seed, and a description that gives text in all four seeds is rewritten.
Before any model run on these pages, we judged whether each photograph shows
the topic of its section, with the help of the assistant named in the LLM Usage
Statement. Then we rewrote the descriptions of the 32 photographs that did not.
None of the models we evaluate took part in this judgment.
The four generated pages of the weight-lens evaluation set (rp-001 to rp-004)
are built in the same way. In an
earlier version of all these pages, each image slot was a CSS gradient
placeholder.
We call the current version the photo pages.

\paragraph{Live-page cells.}
The live-page set has 43 later cells, short of the registered target of 50.
Under the registered rule, a cell enters only when the target section forms a
distinct block on the screenshot and contains at least one element of the DOM
list, and its box is drawn around that block. LLM-based agents drew the boxes
under this rule (LLM Usage Statement of the main paper).
We inspected the drawn boxes on the screenshots by eye. The boxes of the later set were committed before any model output was seen
(Sec.~4.1 of the main paper). In addition, 9 of its 52 candidate cells,
whose target
images had not loaded, were dropped before any prediction.

\paragraph{Decoding and reproducibility.}
Except for the closed models of Appendix~K, generation samples with
temperature $0.7$, top-$p$ $0.95$ and at most 1000 new tokens.
The open models are Qwen/Qwen3-VL-4B-Instruct, Qwen/Qwen3.5-9B,
Qwen/Qwen3-VL-32B-Instruct and
OpenGVLab/InternVL3\_5-8B-HF~\cite{qwen3vl2025,qwen35_9b,internvl35_2025} from
the Hugging Face hub, run with the transformers library in bfloat16.
For Appendix~K, the marked screenshots of the generated-page set, of its
copies without topic names and of the weight-lens evaluation set were sent with
the synthetic persona profiles to the two closed models. These screenshots
include the live Naver page and two live portal screenshots.
The protocol files, the code, the per-trial logs, the generated pages and the other data will be released. The live-page screenshots are copyrighted, so the release gives
their URLs, capture dates and the capture script in their place.

\section{What a Prompt Reaches: Further Tests}
\label{supp:diagnosis}

The main paper does not use the tests of this appendix. They were registered
for the question of Sec.~4.2 of the main paper, on traits other than
interests, and we report them for completeness.

Six perceptual traits show no detectable effect on the scanpath of Qwen3.5-9B
on six pages
(Cohen's $|d|\le.048$ against a preregistered $d>.5$, $q\ge.75$, $n=300$ each).
Asked directly ($n=450$), the personas report their traits as predicted except
on the health attention condition ($p=.58$).
By the preregistered criterion ($d_z>.5$, $q<.05$), prompts reach the two
cognitive traits and neither of the two perceptual traits (largest $d_z$
$.43$, four prompts on six live portals with the 9B and 32B models).

We use Set-of-Mark markers~\cite{som2023} throughout, except where the model is
compared with
or trained on human gaze (Appendices~I and J), and the results are shown for
that format, not for coordinate generation.

\section{Pixel Lens: Detector and Robustness}
\label{supp:lens}

\paragraph{Pixel lens.}
For every topic interest in the profile, the pixel lens adds $0.9$ times the
person's degree of interest to the elements in the three SigLIP2~\cite{siglip2}
blobs that
best match the interest text. The degree is $+1$ for Passionate and $-1$ for
Averse, with the intermediate levels $.5$, $0$ and $-.3$. The sum is clipped
to $[-.9,.9]$, and markers above $.3$ are drawn in red with a 3-px box and
those below $-.3$ in translucent gray. The pixel lens never sees which interest is compared.

The true-section
marker applies the same sum to the true sections. The random marker
emphasizes a random set of elements with the same number and sign. It is
matched over elements and not over the at most 60 markers. In a count added
after data collection, the numbers of
red and of gray markers agree in 362 of the 1{,}440 screenshots per
condition of Table~1 of the main paper, with $15.7$ markers per screenshot for
the random marker and $15.2$ for the pixel lens on average.

\paragraph{Blob detector.}
The blobs come from a dense heat map of google/siglip2-so400m-patch14-384. The
image tower embeds $256\times256$-px windows of the screenshot at a stride of
64 px, and the text tower one query phrase for the interest and up to five
competing phrases. The query phrases are in
\texttt{src/lens.py} of the code, except ``health, wellness,
yoga, meditation'' for health, and the competing phrases are in
\texttt{dense\_heatmap.py}. For every window
the cosine similarity to each phrase is z-scored over the windows of the page,
and the margin is the target z minus the largest competing z. Then the margins
are spread over the pixels of their windows and averaged where they overlap.
Finally, the map is z-scored over the page. The pixels above the 90th
percentile form a mask, and each connected component of the mask with at least
400 px is a blob whose score is its mean map value. The three blobs with the
highest scores are kept. An element lies in a blob when the center of its
box lies inside the bounding box of the blob.

The competing phrases are those
of sports, fashion, finance and food and ``technology gadgets, phones,
software'', and a competing phrase is removed from the list only when it
equals the target phrase character by character. Therefore, the technology
target (``technology, gadgets, AI, software, devices'') keeps the technology
competitor and is scored against a near-synonym. In addition, travel and health
have no competitor of their own. We did not correct this in any run of
this paper. It may explain part of the low recovery on technology.

\paragraph{Page-level and persona-level tests.}
Of the 34 significant entries of Table~1 of the main paper, 32 hold in a test
over the 30 persona means and 25 in a test over the 8 page means. Of the 9 lost
at the page level, 6 are entries of the pixel lens against the base prompt. We
ran these tests because the registered test pools 480 pairs per
entry and treats the 8 pages as fixed.
On one sports page (sp-006) and one fashion page (sp-005) the target section
lies below the viewport, so the target dwell is zero in every trial.

At the page level the pixel lens against the
base prompt keeps only 2 of its 8 (4B and 9B finance), against 6 of its 8 for
the pixel lens against the random marker. Its page-cluster bootstrap intervals
(2000 resamples of the pages) exclude zero in 3 of 9 entries against the base
prompt and in 7 of 9 against the random marker. The true-section marker and the
text lens keep 17 of their 18 entries at the page level.

The random marker
alone changes the pole difference by $-.035$ to $+.004$ against the base
prompt. It is below the base prompt in 7 of the 9 entries and is never
significantly above it (one-sided paired $p\ge.17$).

\paragraph{Pages without topic names.}
On pages without the words that name a section's topic the pixel lens stays
significant in every entry, while the text-lens increment falls in 5 of 9
entries, all at 4B or 9B. The protocol behind Table~1
of the main paper registered a second page set to find the effect of these
words. On copies of the generated pages, section headings and navigation chips
carry neutral labels such as ``Latest'', section links read ``More'' and meta
lines omit the topic, while headlines, snippets, product names and photographs
are unchanged.

Both registered predictions hold: the true-section marker and
the text lens are each significant in 9 of 9 entries (at least 7 of 9
predicted). The pixel lens, registered without a prediction, is significant in
9 of 9 against both the base prompt and the random marker. In the registered
paired comparison with the generated-page set, the pixel-lens increment changes
significantly in 1 of 9 entries, where it rises. On the other hand, the
text-lens increment falls significantly in 5 of 9, including all three at 4B,
and rises significantly in 1.

In an analysis added after data collection, about 70\% of the 32B base effect
is lost without topic names. The base-prompt pole difference on the generated
pages falls in every entry (two-sided paired $p\le.007$). On sports, fashion
and finance it falls at 4B from $.076$, $.050$ and $.019$ to $-.001$, $.006$
and $-.001$, at 9B from $.191$, $.110$ and $.182$ to $.092$, $.052$ and
$.113$, and at 32B from $.296$, $.201$ and $.309$ to $.073$, $.063$ and
$.088$. Fashion and finance are taken over the seven generated pages, since
the live Naver page has no copy. With that page, the fashion and finance
values with topic names are $.039$ and $.019$ at 4B and $.196$ and $.308$ at
32B, as in Sec.~4.5 of the main paper.

What remains is above zero at 9B and
32B on every interest ($p<.001$), and at 4B on fashion only ($p=.007$). The
32B base effect without topic names stays above the 4B one on every interest
(two-sided paired $p<.001$). It lies below the 9B one on sports and finance
($p=.007$ and $.003$) and above it on fashion ($p=.042$).

\paragraph{Other model family and encoder.}
\cref{supptab:lens-transfer} shows that the pixel lens transfers to
InternVL3.5-8B, where it also beats the random marker in all three interests
($+.056$, $+.035$ and $+.078$). With the CLIP ViT-L/14-336~\cite{clip2021}
encoder, run at
4B and 9B only, the recovery is at least $.5$ except on sports at 4B, where
it is $.28$.

\begin{table}[t]
\centering\footnotesize
\setlength{\tabcolsep}{4pt}
\caption{The pixel lens with another model family and another encoder. Rec.\ is
the recovery, the pixel-lens value divided by the true-section value. Each
value is a condition minus the base prompt in the pole difference (477--480
pairs per entry). The InternVL rows come from the pixel-lens study, and the CLIP rows
from their own run with their own true-section marker. Every entry has one-sided $p<.05$, and the InternVL rows also hold after
the Benjamini--Hochberg correction.}
\label{supptab:lens-transfer}
\begin{tabular}{llccc}
\toprule
Setting & Interest & Pixel lens & True section & Rec. \\
\midrule
InternVL3.5-8B & Sports  & $+.062$ & $+.085$ & $.73$ \\
               & Fashion & $+.040$ & $+.067$ & $.59$ \\
               & Finance & $+.059$ & $+.064$ & $.93$ \\
\midrule
4B, CLIP       & Sports  & $+.018$ & $+.065$ & $.28$ \\
               & Fashion & $+.052$ & $+.076$ & $.69$ \\
               & Finance & $+.074$ & $+.083$ & $.89$ \\
\midrule
9B, CLIP       & Sports  & $+.034$ & $+.065$ & $.52$ \\
               & Fashion & $+.052$ & $+.075$ & $.70$ \\
               & Finance & $+.072$ & $+.088$ & $.81$ \\
\bottomrule
\end{tabular}
\end{table}

\paragraph{Further interests.}
On four further interests (technology, cooking, travel and health, eight
generated pages each), the text lens is significant in every entry and the
pixel lens in every entry except technology at 32B. Every significant entry
also holds after the Benjamini--Hochberg correction~\cite{benjamini1995}. On
technology the pixel
lens
recovers only $.09$, $.31$ and $.02$ of the true-section marker at 4B, 9B and
32B, while the true-section marker works ($+.080$, $+.137$ and $+.100$).
Therefore, the weakness lies in finding the region.

\paragraph{Earlier dimension-aware pixel lens.}
The earlier lens was told the interest under test and emphasized only its
region. Appendices~E, F and~G report the runs that used it.

\section{Weight Lens: Training and Adapter Tests}
\label{supp:internalise}

\paragraph{Retrained, lost and excluded adapters.}
Six adapters are retrained copies with the same recipe and seeds, since their
original checkpoints were lost. They are the relaxed-gate 4B adapters of
seeds 0--2, the strong-reward 4B adapters of
seeds 1--2
and the seed-1 4B adapter of a strong-reward variant that sees the true-section marker in training.
The strong-reward 9B adapter of seed 0
was lost and not retrained. Therefore, the strong-reward 9B row of Table~3 of
the main paper averages seeds 1--4. The first-gate adapter is excluded from the
photo-page evaluation, since its checkpoint was overwritten by a later run
with the same name. The variant with the true-section marker in training is
not in Table~3 of the main paper. Over three seeds, its held-out pole
difference is $+.434\pm.077$ on fashion and $+.218\pm.097$ on travel, with a
median of 7 to 12 fixations.

Training does not reproduce bit for bit. Retraining the seed-0 relaxed-gate 4B
adapter moves its held-out pole differences on fashion and travel from $+.352$
and $+.261$ to $+.334$ and $+.204$ on the earlier version of the evaluation
pages (Appendix~A). The original adapter was lost, so this comparison cannot
be repeated on the photo pages.

\paragraph{Training settings.}
Beyond Sec.~3.3 of the main paper, each step takes 2 prompts with 8 rollouts
each (top-$p$ $.95$, at most 700 new tokens) and makes a single
AdamW~\cite{adamw2017} update at
learning rate $10^{-5}$. The KL term is taken against the same model with the
adapter disabled. Every 20 steps its weight (start $0.02$, bounds
$[0.005, 0.5]$) is doubled or halved to hold the KL near the target of $0.15$
nats per token. \texttt{grpo.py} of the code gives the remaining
settings, and the values above are those of the run configurations, which
override its defaults. The reward is set to 0 when the answer does not parse,
names no valid marker or a marker id that does not exist on the page, has fewer
than 5 or more than 40 fixations, repeats a marker in more than $.30$ of the
fixations, or uses fewer than 5 distinct markers.

Training uses the 30 personas of Sec.~4.1 of the main paper and 9 pages
(locally built portals, public shop test sites, one live portal capture and
generated pages). Six of these pages show a fashion section and one a travel
section, but no reward and no training profile uses these interests. None of
the pages is in the weight-lens evaluation set, and
Appendix~H reuses the two locally built portals. There is no validation split.
The KL target follows a pilot adapter that was evaluated on other pages and
personas. The settings above were fixed before any adapter was evaluated on
the evaluation set (see the discarded runs below). In addition, every run is
evaluated at its final step with no checkpoint selection. Otherwise the
evaluation pages
were used only to set the limits of the gate and of its registered relaxation.

\paragraph{Runs discarded before evaluation.}
Two runs of the strong reward were stopped before any evaluation. In the first,
the KL target of $0.02$ was unreachable, the weight doubled up to its cap of
$2.0$, and the reward and the validity rate collapsed. In the second, the KL
estimator had a gradient whose expectation is zero, so the weight had no
effect. Then we switched to another estimator, the target of $0.15$ and a
bounded weight, which every reported run uses.

\paragraph{Gate definition.}
The naturalness gate multiplies the reward by 0 unless three conditions hold.
First, the number of fixations $n$ lies in a range. Second, the entropy
$H=-\sum_k p_k\log_2 p_k$ of the dwell over sections is at least a minimum,
where $p_k$ is the share of the scanpath's dwell on section $k$. Third, the
share $c$ of fixations at the clip limits is at most a maximum.
The clip limits are 80 and 3000\,ms. The prompt asks for dwell times in this
range, we clip every parsed dwell time to it, and it holds 97.6\% of the
351{,}318 fixations recorded in UEyes~\cite{ueyes2023chi}.
The first gate used $15\le n\le 60$, $H\ge1.0$ bits and $c\le.2$. These
limits come from the outputs of the untrained 4B model on the earlier version
of the evaluation pages (Appendix~A), which pass them in 88\% of cases. On the
photo pages the untrained model passes them in 76\% of its outputs.

\paragraph{Relaxed gate.}
The protocol preregistered a single relaxation for the case where the gate
silenced the reward, and the first gate did silence it. On the earlier version
of the evaluation pages the trained model stayed close to the untrained one,
with 40 fixations and $1.93$ bits, and showed no effect on the held-out
interests ($+.007$ and $+.009$, both null). The relaxed gate uses $10\le n\le
60$, $H\ge0.6$ bits and $c\le.3$, with $n$ and $H$ set between
the first gate and the collapsed scanpaths of the strong reward.

The relaxation was registered with three predictions, and 3 of the 5 seeds we
ran meet all of them. First, the held-out fashion and travel
differences exceed those of the base model at $q<.05$ with a size between
$+.1$ and $+.4$. All five seeds are at $q<.001$, but seed 1 exceeds the $+.4$
bound on fashion ($+.527$). Second, the median fixation count stays at 10 or
more, the section entropy at $0.6$ bits or more, and fewer than 60\% of the
Averse-pole scanpaths place no dwell on the target. Seed 2 exceeds the 60\%
bound. Third, the three seeds planned at that time agree in direction.

Every seed of the strong reward and of the relaxed gate, at 4B and at 9B,
exceeds the base model on fashion and travel at $q<.001$. Over its five seeds
the held-out pole difference of the relaxed gate is at least $+.287$ on
fashion and $+.176$ on travel at 4B and at least $+.472$ and $+.541$ at 9B,
above the untrained model with the text lens in Table~3 of the main paper.
The concentration, the dwell share in the most-visited section, is $.51$,
$.58$, $.51$, $.55$ and $.63$ for the five relaxed-gate 4B seeds, within
$.06$ of the $.49$ of the 32B base model in three.
However, under the strong reward at 4B, the median fixation count
falls from 27 to 7--13 over the five seeds, and 56--89\% of the Averse-pole
scanpaths place no
dwell on the target, against 14\% for the base model.

\paragraph{Diversity of the training interests.}
The registered diversity prediction holds on travel but fails on fashion. It
states that the strong reward on the 12 interests and the text-size need
transfers to the held-out interests more than a reward on sports alone. The
paired difference is $+.250$ ($p<.001$) on travel and $-.069$ ($p=.70$) on
fashion, where the sports-only reward already transfers (pole differences of
$+.776$ on fashion and $+.200$ on travel).

\paragraph{Training with the text lens.}
Training with the note improved on the original adapters significantly in 1 of
18 entries. We retrained the 4B strong-reward, 4B
relaxed-gate and 9B relaxed-gate settings with the note prepended to the
training prompt (three seeds each) and registered three comparisons. With the
text lens, the new adapters are above the original ones in 3 of 18 entries,
significantly in 1. With the text lens they are above the untrained model in
12 of 12 entries at 4B and in 0 of 6 at 9B, where all 6 differences are
negative. With the base prompt they are weaker than the original ones in 18 of
18 entries (mean difference $-.333$).

\paragraph{Irrelevant attributes and the adapters.}
The relaxed-gate 9B adapter separates personas on 3 of the 5 attributes that
should not move gaze in every seed, and the 4B adapters less often. We evaluate all five
training seeds, as registered, on personas that differ only in Korean food
preference, AI for documents, sleep schedule, rock music or knitting
(held-out pages, 40 pairs per test). The relaxed-gate 9B adapter separates
them on Korean food, rock music and knitting in every seed ($q<.05$), at
$.76\pm.16$, $.41\pm.16$ and $.75\pm.11$ of the sports effect (the ratio of
Appendix~J), and with the
pixel lens on top at $.67\pm.08$, $.43\pm.15$ and $.67\pm.11$, again in every
seed.

The 4B adapters separate on Korean food, which some pages cover in a food
section, in 4 of 5 seeds (strong reward, $.22\pm.09$) and 3 of 5 (relaxed gate,
$.22\pm.07$). The relaxed gate separates on rock music and on knitting in 3 of
5 seeds each (at most $.29$ of its sports effect), and the strong reward in 1
of 5 seeds each. However, the untrained 4B and 9B models separate on none of
the five attributes (smallest $q=.084$).
The registered floor check of Appendix~J covers the base prompt and the text
lens on the untrained models only, so its floor (b) is not measured for the
adapters.

\paragraph{Pixel lens on the adapters.}
The pixel lens is weaker on two of the three adapters. It is significant in
3 of 6 interests on the untrained 4B model and 4 of 6 on the untrained 9B
model. However, on average over five seeds it is significant in 1.6, 3.4 and
1.6 of 6 interests on the strong-reward 4B, relaxed-gate 4B and relaxed-gate 9B
adapters. On the relaxed-gate 9B adapter it lowers sports in 4 of 5 seeds (by
$.008$ to $.063$) and raises it in the other ($+.057$).

\section{Scale: Further Results}
\label{supp:scale}

\paragraph{Earlier run of the scale comparison.}
The rise with model size first appeared in one registered protocol run
unchanged at 4B, 9B and 32B on the earlier version of the generated-page set,
with the earlier dimension-aware pixel lens (14{,}400 trials per model). In
that run the mean base-prompt pole difference over the 24 page-by-interest
entries, which lie on 11 pages, rose from $.044$ at 4B to $.164$ at 9B and
$.261$ at 32B. The scale trend in the main paper uses the pixel-lens run on the
photo pages, which gives $.044$, $.156$ and $.267$. It gives a positive
difference on 11 of 11 pages at 32B against 10 of 11 at 4B, and the
same order of the three models on each interest. However, on the copies of
these pages without topic names, the 32B value stays above the 4B one but
falls below the 9B one on sports and finance (analysis added after data
collection, Appendix~C).

\paragraph{Results on the weight-lens evaluation set and on live pages.}
On the weight-lens evaluation set the 32B base model exceeds the 4B base by
$+.300$ on fashion and $+.181$ on travel ($q<.001$ each, paired
Wilcoxon~\cite{wilcoxon1945} with Benjamini--Hochberg correction over the two
held-out interests). The set is four
generated pages and two live portal pages over 2{,}040 trials per model, 840 of
them with the base prompt. The 32B base model answers with a median of 25
fixations there, against 27 at 4B. On the two portal pages the difference is
$+.019$ at 4B (not significant, $n=80$, $p=.204$), $+.098$ at 9B and
$+.225$ at 32B.

\paragraph{Effect of scale on text size.}
The registered prediction that text size, a secondary perceptual check,
stays non-significant at scale ($p \ge .05$ and $|\Delta| < .05$, registered
before we read the 32B run) fails at 32B, and the effect there does not
replicate. The pole difference is $-.004$ at 4B ($p=.745$), $-.017$ at 9B
($p=.493$) and $+.030$ at 32B ($p=.048$). Two further registered runs at 32B,
each with a new pair of generation seeds, give $+.010$ ($p=.715$) and $+.036$
($p=.211$). However, the sign is positive in all three runs.

\section{Attention: Measurement, Masks and\texorpdfstring{\\}{ }Registered Predictions}
\label{supp:mechanism}

\paragraph{Measurement.}
In each layer the excess attention is the share of the generated tokens'
attention to image tokens that falls inside the target section, averaged over
heads and generated positions, minus the area share of the target (its
fraction of image tokens). The late layers are layers 28--36 of 36 at 4B and
those at the same relative depth at 9B and 32B (the 64 layers of 32B are
interpolated to 36 points). The target attention of Sec.~5 of the main paper is
the pole difference in this excess over the late layers.

We chose the late layers because at 4B on the earlier version of the generated pages
the earlier dimension-aware pixel lens raised the late-layer pole difference
in excess attention from $+.006$ to $+.023$ on fashion and from $-.039$ to
$+.156$ on travel (seed 0, Appendix~A). However, on the photo pages it raises
the fashion difference ($+.009$ to $+.031$) but lowers the travel one ($+.052$
to $-.006$). At 9B only the 8
full-attention layers return attention weights and enter the measurement and
the masks. Therefore, the last-quarter mask blocks layers 28 and 32 of 32 and
leaves
the 6 linear-attention layers of that quarter unmasked. One travel page,
whose target lies outside the screen, is excluded from the attention runs of
this appendix.

\paragraph{Weight lens and scale.}
Only at 9B do the adapters raise the target attention on both interests
(\cref{supptab:attn-models}, the values behind the comparison across models in
Sec.~5
of the main paper).

\begin{table}[t]
\centering\footnotesize
\setlength{\tabcolsep}{3pt}
\caption{Target attention of the base models and the adapters on five pages of the weight-lens evaluation set, averaged over three generation seeds for the base models, three adapters at 4B and two generation seeds of one adapter at 9B (training seed 0 for the relaxed gate, seed 1 for the strong reward, whose seed-0 adapter was lost). Dwell $\Delta$ vs.\ base: paired change in the held-out pole difference (fashion/travel) against the base model of the same size (the 32B row against the 4B base), on the whole weight-lens evaluation set for one adapter per row (training seed 0, seed 1 for the 9B strong reward), unlike Figure~6 of the main paper, which averages three adapters over the five pages. ---: the reference itself.}
\label{supptab:attn-models}
\begin{tabular}{@{}lccc@{}}
\toprule
 & \multicolumn{2}{c}{Target attention} & \\
\cmidrule(lr){2-3}
Condition & Fash. & Trav. & Dwell $\Delta$ vs.\ base \\
\midrule
Base 4B              & $+.010$ & $+.055$ & --- \\
Strong reward        & $+.023$ & $+.010$ & $+.520$/$+.420$ \\
Relaxed gate         & $+.038$ & $+.029$ & $+.221$/$+.145$ \\
\midrule
Base 9B              & $+.029$ & $+.018$ & --- \\
\quad relaxed gate   & $+.103$ & $+.068$ & $+.361$/$+.400$ \\
\quad strong reward  & $+.170$ & $+.128$ & $+.704$/$+.524$ \\
\midrule
Base 32B             & $+.026$ & $+.009$ & $+.300$/$+.181$ \\
\bottomrule
\end{tabular}
\end{table}

\paragraph{Intervening on attention.}
By the registered bound of $.5$ on the mean retention, the late layers count
as the channel of the true-section marker at 32B ($.11$) but not at 4B or 9B,
where it keeps $.96$ and $.52$ of its travel increment and its fashion
increments are too small for a ratio.
\cref{supptab:mechanism} gives the retention, and the upper block of
\cref{supptab:words-attn} gives the late-layer increments of the lenses. The
mask leaves the prompt encoding and
the vision encoder untouched, sets the target share in the masked layers to
exactly zero and keeps the output intact. Under the
late-layer mask (seed 0) the marker keeps $3.49$ and $.52$ of
its fashion and travel dwell increments at 9B and $-.64$ and $.86$ at 32B,
and the pixel lens keeps $2.31$ and $1.26$ at 4B. However, at 9B the fashion
increment of the marker is too small for a ratio (at most $.03$). At 32B it is
small as well ($+.029$ pooled over five seeds, \cref{supptab:words-attn}), so
the mean of $.11$ rests on the fashion ratio of $-.64$, and on travel alone the
marker keeps $.86$. The text lens
keeps a mean of $.66$ under the late-layer mask and $.46$ under the full
mask.

\begin{table*}[t]
\centering\footnotesize
\setlength{\tabcolsep}{4pt}
\caption{Retention of the pole difference when the attention to the target is
masked in one quarter of the layers or in every layer, on five pages of the
weight-lens evaluation set with generation seed 0. The unmasked rows give the
pole
difference (the increment over the base prompt for the true-section marker
and the text lens) from these masking runs, so they differ from Table~3 of
the main paper. Pairs give fashion/travel. For the 4B adapters (Strong:
strong reward, Gate: relaxed gate) the retention is the mean over the two
interests. The fashion increment of the marker ($+.028$) is too small for its
ratios to be read. ---: a quarter we did not mask.}
\label{supptab:mechanism}
\begin{tabular}{@{}lccccc@{}}
\toprule
 & \multicolumn{4}{c}{4B} & 32B \\
\cmidrule(lr){2-5}\cmidrule(l){6-6}
Masked layers & True-section marker & Text lens & Strong & Gate & Base \\
\midrule
Unmasked, fash.\ & $+.028$ & $+.129$ & $+.589$ & $+.240$ & $+.388$ \\
Unmasked, trav.\ & $+.043$ & $+.211$ & $+.586$ & $+.210$ & $+.273$ \\
\midrule
First quarter     & $1.66$/$2.41$ & ---          & $.97$ & $1.07$ & --- \\
Second quarter    & $-.20$/$2.06$ & ---          & $.93$  & $.93$ & --- \\
Third quarter     & $.09$/$1.75$ & ---          & $.67$  & $.66$  & $.79$/$1.02$ \\
Last quarter      & $2.29$/$.96$ & $.86$/$.46$  & $.65$  & $.70$  & $.79$/$.90$ \\
Every layer       & $.87$/$.72$ & $.56$/$.35$  & $.36$  & $.07$  & $.62$/$.55$ \\
\bottomrule
\end{tabular}
\end{table*}

\paragraph{What was preregistered.}
Of the 12 registered predictions on attention counted in Sec.~5 of the main
paper, 6 held and 6 failed. They form the first set of registrations
on attention (weight lens, scale and the earlier lens). Two of them also stated
that the target attention would not correlate with dwell, and that part is not
judged. A registered noise check discards, as
noise, any correlation between the two pole differences over the 48 to 50 pairs
if the untrained 4B model reaches $|\rho|\ge.3$ at $p<.05$ in any seed, and its
seed 1 reaches $\rho=+.34$ ($p=.016$) on the photo pages. The six predictions
that held:
\begin{enumerate}[leftmargin=*,nosep]
\item At 32B the increment of the earlier lens lies in the middle layers,
  not the last quarter (two seeds).
\item Registered after prediction 8 failed: the late-layer rise of the
  strong-reward 9B adapter over the base model is at least $+.08$ in both
  generation seeds.
\item At 4B the late layers are not the channel of the earlier lens
  (retention $2.41$ under the late-layer mask, $1.62$ in the rerun with the
  true-section marker, against the registered $.5$).
\item The two 4B adapters keep at least $.5$ of their pole difference under the
late-layer mask ($.65$ and $.70$).
\item For both adapters the smallest retention over the first three quarters
  lies in the middle half (third quarter, $.67$ and $.66$).
\item The 32B base model keeps at least $.79$ of its pole difference under the
  third-quarter and the late-layer mask.
\end{enumerate}
The six predictions that failed:
\begin{enumerate}[leftmargin=*,nosep,resume]
\item The target attention of the strong reward at 4B stays near zero.
  It is $+.030$/$+.015$ (fashion/travel) in seed 1 and $+.015$/$+.006$ in
  seed 2 (all $p<.001$).
\item The 9B increase stays within $+.08$. The strong reward exceeds it
  ($+.141$ and $+.110$), while the gate stays within it ($+.074$, $+.051$).
\item The pixel lens re-allocates attention at 32B. The earlier lens reaches
  $+.035$ against $+.10$.
\item Masking every layer removes the increment of the earlier lens (positive
  control of the mask). It keeps $.55$ with $p=.062$, against a retention
  below $.5$ or $p<.05$.
\item At 32B the third-quarter mask lowers the increment of the earlier lens
  more than the late-layer mask. It keeps $.59$/$.92$ of the fashion/travel
  increments ($+.054$/$+.212$) against $.04$/$.59$.
\item For both adapters the smallest retention over the first three quarters
  lies below the last-quarter retention. The strong reward keeps $.67$ in the
  third quarter against $.65$ in the last.
\end{enumerate}
The other registered tests on attention are not among the 12.
Those of the earlier lens on further pages are not reported here, and
Appendix~J reports the later registrations on the pixel lens, the text lens
and the marker-naming control, and the re-measurement on the generated-page
set. The
re-measurement held one further prediction, that the text lens changes the
target attention by at most $.02$ in each of the 9 entries (largest $+.011$).

\section{Live-Page Evaluation: Cells and Controls}
\label{supp:realpages}

\paragraph{Cells and conditions of the live-page evaluation.}
A cell succeeds
when the condition minus the base prompt is positive at one-sided paired
Wilcoxon $p<.05$ with at least five pairs.
The 87 cells consist of an \emph{earlier set} of 44 cells from the earlier
round described below and a \emph{later set} of 43 cells. As registered, we
report the later set both on its own and together with the earlier set. The
earlier
set lies on 8
sites (MSN editions counted as one), with 41 cells on sports, fashion or
finance. The later set lies on 13 sites that share only MSN and BBC with it,
with 8 cells on shopping sites and 18 cells on technology, cooking or travel
against 3 in the earlier set. Together the 87 cells lie on 61 screenshots of
19 sites.

\paragraph{Pixel lens on top of the text lens.}
The text lens with the pixel lens on top beats the text lens alone in 28,
18 and 21 cells at 4B, 9B and 32B and loses in 6, 5 and 11. It wins in 17, 6
and 16 of the 44 earlier cells, where the notes are the same, and in 11, 12
and 5 of the 43 later cells. However, on the later cells the combined condition
ran separately and wrote its note again, which equals the note of the text-lens
run in only 34
to 37\% of the trials (Appendix~J).

\paragraph{Technology and shopping cells.}
The pixel lens succeeds in few technology and shopping cells. On the 10 later
technology cells it beats the base prompt in 4, 2 and 2 at 4B, 9B and 32B,
against 7, 8 and 9 for the true-section marker. Thus, the blob detector misses
technology regions, as on generated pages (recovery at most $.31$, Sec.~4.3
of the main paper).

The eight shopping cells lie on four screenshots from two
sites (seven on one Amazon front page), and they ran with the news-reading
task sentence of Appendix~A. A trial with an empty scanpath has no dwell share
(Appendix~A), so at 32B a test keeps only 5 to 41 of the 60 pairs on the
Amazon cells. In a recount added after data collection that scores an empty
scanpath as zero dwell share, the pixel lens beats both the base prompt and the
random marker in 1, 1 and 2 of the 8 shopping cells. In addition, no
count of the text lens or the true-section marker changes by more than 2 cells.

\paragraph{Random marker on all cells.}
The random marker alone beats the base prompt in 8, 8 and 4 of the 87 cells
at 4B, 9B and 32B, where the uncorrected threshold of $.05$ gives $4.4$ by
chance. A registered addition ran the random marker on the 44 earlier cells
with the same random selection as on the later cells, reusing their
base-prompt and pixel-lens runs. Over all 87 cells the pixel lens
beats the random marker in 34, 31 and 42 cells. It beats both the base prompt
and the random marker in 26, 21 and 27 (14, 12 and 13 of the later and
12, 9 and 14 of the earlier cells).

\paragraph{Agent re-annotation.}
Target boxes redrawn by a separate LLM-based annotation agent, which did not
see the first boxes, change no count by more than one cell. The agent redrew
the target section in all 43 later cells and 20 random earlier cells. It agrees
that the target section is distinct in 60 of these 63 cells, with a
median IoU
of $.99$ against the first boxes on the later set. The comparison re-scores
these 60 cells with both sets of boxes. In the other 3 earlier cells the agent found no
distinct target section, and 24 earlier cells were not redrawn.

\paragraph{Earlier round.}
In the earlier round, a registered rule for which cells fail was rejected,
and a registered intervention fell short of its threshold. The round used the
earlier dimension-aware pixel lens of Appendices~C and~F at 9B, on 26 cells we
had collected and 18
cells registered in advance, where that lens succeeds in 8 and 1 cells.
The registered rule predicts a success when the target section carries at
least one marker, at least four marked elements lie in the three best blobs
and at least half of them belong to the target section. On the 18 held-out
cells it was right in 67\% of the cells, against a threshold of 75\%.
An intervention that spends the marker budget inside the highlighted region
first raised the successes from 8 to 13 of 26 cells, short of the registered
threshold of 14 (McNemar~\cite{mcnemar1947} $p=.227$ on 11 discordant pairs).

\section{Personas with Two Interests}
\label{app:cross}

We crossed two interests at the five levels of the corpus, from Passionate to
Averse, with every other attribute fixed, sports and finance on eight pages and
technology and finance on six (7{,}000 scanpaths per model, no pixel lens).
Two of these pages are the locally built portals on which the adapters were
trained (Appendix~D).
Each interest moves the dwell share on its own section in all five models
(three base models and two 4B adapters) and both pairs ($20$ of
$20$, $q<10^{-5}$). The registered bound states
that the first interest moves the second section by at most one third of its
own effect. It holds for technology and finance in 4 of 5 models and fails for
sports
and finance in 4 of 5, where the change is negative.

\section{Joint Training with Human Gaze}
\label{app:human}

\paragraph{Data and model.}
We fine-tune a rank-64 LoRA~\cite{lora2021} on the language model and the
vision--language
connector with the raw eye-tracker logs of UEyes~\cite{ueyes2023chi}, 7 seconds
of free viewing per image (official split, 1{,}872 training and 108 test
images). UEyes has no element boxes. Therefore, the gaze models of this
appendix answer with coordinates
(\texttt{[[x, y, ms], ...]}). A human row targets the sequence of one viewer
with at most 30 fixations, with ``an average web user'' in the persona slot.

The joint model trains a fresh adapter from the base model for one epoch on
these rows and on 3{,}780 scanpaths that the relaxed-gate 4B model
(Sec.~3.3 of the main paper) produces with real persona text.
These scanpaths come from the training pages, the 12 training interests and the
training personas only. Therefore, the evaluation pages, the held-out interests
and the evaluation personas never reach the model. The rounded dwells of the
teacher (median 800\,ms) are mapped by quantiles onto the dwell distribution
of the UEyes training viewers.

\paragraph{Registered criteria and their amendments.}
Three criteria decide success. One of them was added after the first stage, the
training of the human-gaze model, and one further registered prediction was
replaced. We evaluate the 840 scanpaths of the weight-lens evaluation set
without a pixel lens. A
fixation counts for the section whose box contains it, and the persona effect
is the Passionate minus Averse difference in target-section dwell share,
where a cell is one interest, page, persona and seed. (i) On fashion and
travel, the persona effect exceeds that of the human-gaze model of the same
size, which is trained on UEyes alone (paired one-sided Wilcoxon over cells,
Benjamini--Hochberg $q<.05$). (ii) The ScanMatch~\cite{scanmatch2010} to the
viewers of the same image on the 27 web test images is at least $0.9$ times
that of the human-gaze model. (iii) The image-specific share, this ScanMatch
minus that to the viewers of eight other images of the same type, has a 95\%
bootstrap interval above zero. The human ScanMatch in
\cref{supptab:human} is the mean over all pairs of viewers of an image.

We added (iii) after the first stage, because
the registered first-stage ScanMatch threshold of $.30$ lies below the $.328$
of randomly
paired humans and is met by a human-like path shape alone.
That threshold was kept and is met (the human-gaze model reaches $.433$), and
(ii) was registered together with (i).

We had also registered that the
human-gaze model keeps at least half of the persona effect of the untrained
model. This ratio could not be judged at the time, because in the coordinate
format the untrained model has almost no persona effect (six-interest sum
$+.013$ on the earlier version of the pages, on which we amended the rule, and
$+.095$ on the photo pages, where no interest is significant). After seeing
this and before scoring the human-gaze model, we replaced the
prediction with the test that
its persona effect exceeds zero. The test fails (six-interest sum $-.095$,
$q\ge.40$ on every interest).
On the photo pages the registered ratio can be computed, and it fails as
well (the ratio of the two sums is about $-1$).

\begin{table*}[t]
\centering\footnotesize
\setlength{\tabcolsep}{4pt}
\caption{Joint training on human and persona-conditioned scanpaths. Six-int.\
sum is the persona effect summed over the four trained and two held-out
interests. Fashion and travel give the paired difference in persona effect
against the human-gaze model of the same size (ref.) on the same cells, and
every entry is significant ($q<.05$ in every seed). ScanMatch and the
image-specific share (as a ratio to the human share of $+.075$) are measured
on the 27 UEyes web test images. Dwell is the median fixation dwell of the
persona-conditioned scanpaths on the weight-lens evaluation set, and of the
UEyes
fixations for the human viewers. Multi-seed rows give the mean and the
standard deviation over training seeds. --- marks an entry that does not apply
or was not measured. $^{*}$Training seed 0.}
\label{supptab:human}
\begin{tabular}{llcccccc}
\toprule
Model & Seeds & Six-int.\ sum & Fashion & Travel & ScanMatch & Image-specific & Dwell (ms) \\
\midrule
Human viewers (UEyes)        & ---  & ---                 & ---           & ---           & $.427$          & $1.00$          & 279 \\
Base 4B, no training         & 1    & $+.095$             & ---           & ---           & $.139$          & $.01$           & --- \\
\midrule
Human-gaze model, 4B         & 1    & $-.095$             & ref.          & ref.          & $.433$          & $1.08$          & 297 \\
Joint model, 4B              & 3    & $+1.766\pm.129$     & $+.329\pm.036$ & $+.201\pm.027$ & $.433\pm.007$ & $1.00\pm.09$    & 263 \\
Human-gaze model, 9B         & 1    & $+.028$             & ref.          & ref.          & $.433$          & $.98$           & --- \\
Joint model, 9B              & 3    & $+1.810\pm.170$     & $+.207\pm.038$ & $+.219\pm.014$ & $.430\pm.005$ & $.93\pm.09$     & 262$^{*}$ \\
Human-gaze model, 32B        & 1    & $-.030$             & ref.          & ref.          & $.439$          & $1.14$          & --- \\
Joint model, 32B             & 1    & $+2.110$            & $+.289$       & $+.263$       & $.425$          & $.92$           & 263 \\
\bottomrule
\end{tabular}
\end{table*}

\paragraph{Main result.}
\cref{supptab:human} shows that the joint model meets all three criteria at 4B,
9B and 32B and in each training seed at 4B and 9B.
A second training seed at 32B, evaluated only on the earlier version of the
evaluation pages (Appendix~A), also meets them ($+2.061$, ScanMatch $.428$).
Four new evaluation seeds at 4B give the same result
($+1.595$, all six interests $q<.05$). In addition, the joint model passes all
criteria with the teacher rows at about 10\%, 18\% and 40\% of the training
rows, and with a teacher trained with the strong reward.

\paragraph{Reward training on top of the human-gaze model.}
Reward training on top of the human-gaze model meets all criteria in 2 of 20
runs. The runs use the gated reward of Sec.~3.3 of the main paper in coordinate
form, with the human-gaze model as the KL reference, at 4B and 9B. The 16
runs of 500 steps fail. The 4 runs of 1{,}500 steps all meet (i) and (iii),
and 2 of them, the first 4B seed and the 9B run, reach the ScanMatch threshold
of (ii).

We found these two runs late. The first evaluation of the 1{,}500-step 4B run
had used an adapter restored from step 249, so we did not pursue the setting.
The other 3 runs were evaluated only after we corrected the restored adapter.
With the adapter of the human-gaze model scaled down, the reward fails (ii),
and on the
base 4B model it separates the held-out interests in 1 of 3 seeds without
human-likeness and fails at 9B. We stopped four further seeds of settings
that had already failed, and they have no result.

\section{Text Lens: Variants, Checks and\texorpdfstring{\\}{ }Comparison with the Pixel Lens}
\label{supp:words}

\paragraph{Prompt conditions.}
The \emph{three-step prompt} adds a plan between the note of the text lens and
the gaze prompt. In the second step the model receives the plain-marker
screenshot, the persona and its own note and plans in two or three sentences
where this person's eyes will go, without marker numbers. The plan is decoded
greedily (temperature 0) once for each page, persona and pole, and the third
step prepends the note and the plan to the user turn of the gaze prompt.

The
\emph{text ceiling} prepends a sentence that we wrote from the pole and the
interest under test (``This person looks for X content first.'' or ``This
person avoids X content.''). The \emph{marker-naming control} takes the
selection of the pixel lens, the markers with a score above $.3$ and below
$-.3$, and states their numbers at the start of the user turn (``Note: on this
page, the elements most relevant to this person are marked 2, 24, 25; the
elements this person would avoid are marked 7, 8.''). Its screenshot carries
plain markers, as under the base prompt. \texttt{run\_p16.py} of the code holds the full prompts.

\paragraph{Generated pages.}
On the generated-page set the three-step prompt beats the text lens in 9 of 9
tests ($p<.01$ in every test). In a registered addition that reuses the notes
and plans of the three-step run in every trial, the pixel lens on top of it
adds in 7 of 9 tests. A first run of this addition read notes left by other
runs, which equal those of the three-step run in only 18 to 54\% of the
trials. We report it as a deviation and judge the addition on the rerun, and
the first run also gives 7 of 9. On the 43 later
live cells the note of the pixel lens on top of the text lens was generated
again. It equals the note of the text-lens run in only 34 to 37\% of the
trials, against all trials on the 44 earlier cells.

\paragraph{Text ceiling and InternVL.}
The text lens adds more than the text ceiling at 4B and less at 32B. In the
text-ceiling run on the generated-page set (sports, fashion, finance), the text
lens adds $.166$--$.280$, $.116$--$.149$ and $.120$--$.158$ to the pole
difference at 4B, 9B and 32B, and the text ceiling $.104$--$.212$,
$.107$--$.138$ and $.185$--$.258$. Thus, the text lens adds $1.26$--$1.59$,
$.99$--$1.19$ and
$.61$--$.66$ times as much as the ceiling.

On top of either text prompt the pixel lens adds at 4B and 9B, and at 32B on
at most 1 of 3 interests. The pixel lens on top of the text lens was run twice
on the Qwen models (with the pixel-lens study and with the text-ceiling run,
which regenerate the notes). Its increment is significant (one-sided paired
$p<.05$) on 3 of 3 interests at
4B and 9B in both runs and on 1 and 0 of 3 at 32B. On top of the text ceiling
it adds $+.015$ to $+.085$ at 4B and 9B, significant in all 6 tests and in 3
of them over the 8 page means (4B fashion and finance, 9B finance). At 32B it
is significant on 1 of 3 interests (finance, $+.019$).

On InternVL3.5-8B,
in a separate run with its own base prompt, the text lens adds $+.072$,
$+.049$ and $+.063$ on sports, fashion and finance, the pixel lens $+.052$,
$+.042$ and $+.057$, and the pixel lens on top of the text lens $+.074$,
$+.034$ and $+.045$ (one-sided $p<.001$ in every test). Its pixel-lens values
differ slightly from \cref{supptab:lens-transfer}, a separate run.

\paragraph{Attention on the target.}
By the registered rule, the text lens moves the target attention of Sec.~5 of
the main paper in 4 of 6 entries and the pixel lens in 3 of 6. However, every
increment of the text lens is $+.010$ or less. The upper block of
\cref{supptab:words-attn} gives the increments on five pages of the
weight-lens evaluation set of Sec.~4.4 of the main paper, fashion on two and travel on three. We had
registered that a condition moves the target attention when at least 4 of 5
seeds rise and the pooled test gives $p<.05$. The text lens moves it on travel
in every model and on fashion at 32B, and the pixel lens on fashion at 4B and
32B and on travel at 9B.

Computed after the data, at 4B the travel values of all four conditions come
from one of the three travel pages (rp-001). On the other two the target
attention stays within $.011$ of zero in all four conditions. Without that page
the pixel lens and the true-section marker raise the travel difference in all
five
seeds, by $+.008$ each.
The registered sentence does not say whether the two interests are judged
separately. Judged together, the text lens moves the
target attention in every model ($+.006$, $+.007$ and $+.005$, in 4, 5 and 5
of 5 seeds), and the pixel lens only at 9B ($+.011$, 5 of 5 seeds).

\begin{figure}[t]
\centering
\includegraphics[width=\linewidth]{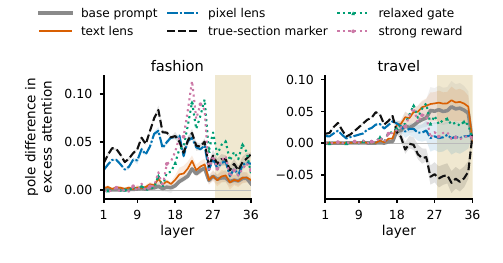}
\caption{Pole difference in excess attention on the target by layer at 4B, on two fashion and three travel pages of the weight-lens evaluation set. Lines are means over five generation seeds (bands give their range) or, dotted, over three adapters. The shaded band marks the late layers. One travel page gives most of the late-layer travel values. The view by layer is post hoc.}
\label{suppfig:attn-layers-five}
\end{figure}

Post hoc, by quarter of the layers, the pixel lens and the true-section marker
change the 4B difference by $3.2$ to $12.9$ standard deviations across pairs
under the base prompt in the first two quarters, against at most $.59$ in any
quarter for the text lens. \cref{suppfig:attn-layers-five} and
\cref{supptab:attn-quarters} give the values, and the ratios are taken before
rounding. The gray bands of the dwell panels of Figure~6 of the main paper span
$\pm.084$ and $\pm.117$ on fashion and travel, one standard deviation across
pairs under the base prompt. The Dwell entries of the true-section
marker in
\cref{supptab:words-attn} pool 97 of 100 pairs on fashion at 4B and 145 of 150
on travel at 32B. Over all valid pairs of the marker these two entries are
$+.048$ and $+.120$, and Figure~6 of the main paper shows the former.

\begin{table}[t]
\centering\footnotesize
\setlength{\tabcolsep}{4pt}
\caption{Change over the base prompt at 4B in the pole difference of excess
attention on the target, by quarter of the layers, on five pages of the
weight-lens evaluation set. The last column (layers 28--36) is the target
attention of the main paper. SD is the standard deviation of the same
difference across pairs under the base prompt, and its value in the last column
gives the gray bands of the attention panels of Figure~6 of the main paper.
Under the pixel lens and the true-section marker the first two columns rise in
all five seeds on both interests. The text lens, the pixel lens and the
true-section marker give paired increments pooled over five generation seeds,
and the adapters give the mean of three
adapters minus that of the untrained model. Computed after the data.}
\label{supptab:attn-quarters}
\begin{tabular}{@{}llcccc@{}}
\toprule
 & & \multicolumn{4}{c}{Layers} \\
\cmidrule(lr){3-6}
Interest & Condition & 1--9 & 10--18 & 19--27 & 28--36 \\
\midrule
Fashion & Text lens & $+.002$ & $+.004$ & $+.005$ & $+.001$ \\
 & Pixel lens & $+.027$ & $+.047$ & $+.027$ & $+.013$ \\
 & True-section marker & $+.036$ & $+.059$ & $+.029$ & $+.018$ \\
 & Relaxed gate & $-.000$ & $+.007$ & $+.048$ & $+.028$ \\
 & Strong reward & $-.001$ & $+.012$ & $+.051$ & $+.013$ \\
 & SD, base prompt & $.0028$ & $.0077$ & $.0220$ & $.0255$ \\
\midrule
Travel & Text lens & $+.000$ & $+.003$ & $+.015$ & $+.010$ \\
 & Pixel lens & $+.012$ & $+.023$ & $-.017$ & $-.037$ \\
 & True-section marker & $+.020$ & $+.033$ & $-.053$ & $-.094$ \\
 & Relaxed gate & $+.000$ & $+.005$ & $+.003$ & $-.026$ \\
 & Strong reward & $+.001$ & $+.005$ & $-.008$ & $-.045$ \\
 & SD, base prompt & $.0019$ & $.0073$ & $.1128$ & $.1628$ \\
\bottomrule
\end{tabular}
\end{table}

\begin{figure*}[t]
\centering
\includegraphics[width=.86\linewidth]{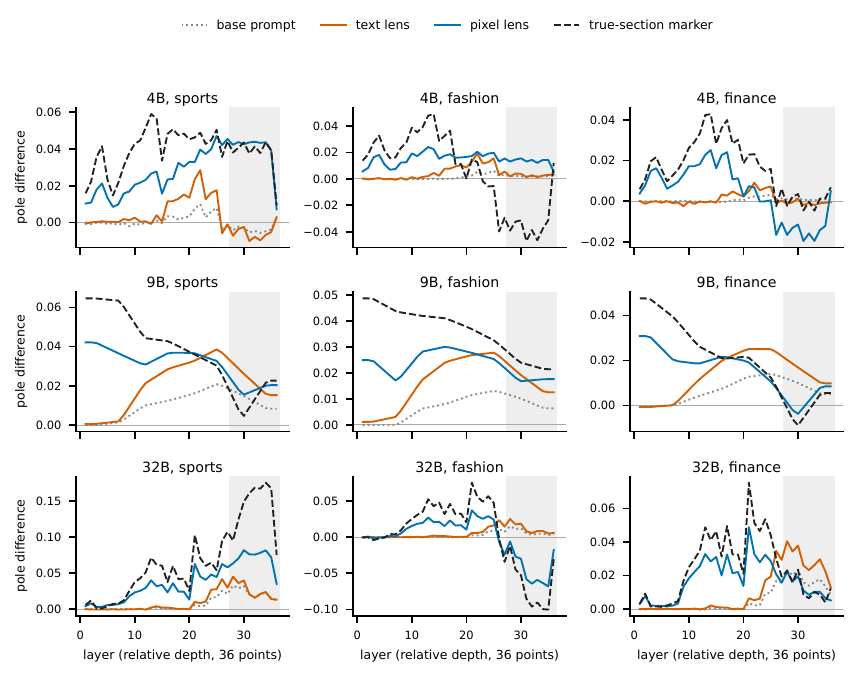}
\caption{Pole difference in excess attention on the target by layer on the
generated-page set, for each model and interest. Lines are means over the pages
whose target is on the screen, personas and five generation seeds. The 64
layers of 32B and the 8 measured layers of 9B are placed on 36 points by
relative depth. The shaded band marks the late layers. The registration lists
this view as a report and sets no criterion on it.}
\label{suppfig:attn-layers-t1}
\end{figure*}

\paragraph{Attention on the generated-page set.}
In the registered re-measurement on the generated-page set, no condition meets
the registered criterion for raising the target attention, a
rise in at least 7 of 9 entries and a fall in none. Registered before data
collection, it covers 3 interests with 8 pages each, the first 10 of the 30
personas and 5 generation seeds in each model for the base prompt, the text and
pixel lenses and the true-section marker. One sports and one fashion page are
excluded because their
target lies off the screen. Each entry is judged by the registered rule stated
above (at least 4 of 5 seeds and a pooled $p<.05$), mirrored for a fall (lower
block of
\cref{supptab:words-attn}).

The pixel lens raises the target attention in 5 of 9 entries, on sports
in every model and on fashion at 4B and 9B, and lowers it in none. In the other
4 the difference falls in all five seeds, but the pooled test does not confirm
a fall, so we state no direction. The text lens raises it in 5 of 9, with a
maximum of $+.011$ within the registered bound of $.02$, while it moves dwell
by $+.118$ to $+.261$. The true-section marker raises it in 4 of 9 and lowers
it in 2, on finance at 9B and on fashion at 32B.

Every entry judged to rise or
fall has $q<.01$ (Benjamini--Hochberg over the 9 entries of a condition). The
dwell increments have the sign of Table~1 of the main paper in 26 of 27
cases. On fashion, the one interest in both blocks of
\cref{supptab:words-attn}, the attention increment changes sign for the
true-section marker in every model and for the pixel lens
at 9B and 32B.

\paragraph{Unregistered checks of these runs.}
In checks that were not registered, single pages carry several verdicts. The
pooled test treats the pages as fixed and counts the five seeds as separate
pairs. Removing one page changes the verdict in 14 of the 27 tests,
including 2 of the 5 in which the pixel lens raises the difference (sports at
9B and 32B). Of these 5 entries, 4 keep $p<.05$ over the 10 persona means and
none over the page means, against 5 and 4 of the 5 entries of the text lens.

By quarter of the layers, the pixel lens and the true-section marker raise the
difference in 9 of 9 entries in each of the first two quarters
(\cref{suppfig:attn-layers-t1}). Under the pixel lens the excess attention on
the target falls for both poles in 6 of 9 entries, and in 3 of them the
difference still widens because the Averse pole falls further. Therefore, a
wider difference does not show that the model attends to the target more.

\begin{table*}[t]
\centering\footnotesize
\setlength{\tabcolsep}{4pt}
\caption{Target attention and dwell. Attn.: increment of the target attention
over the base prompt, pooled over five seeds (upper block: 100 pairs on
fashion, 150 on travel), with the number of seeds in which it rises. $p$:
pooled one-sided paired Wilcoxon test for a rise under the pixel lens. It is
below $.05$ on travel at 4B and 32B although the increment is negative, and
there the seed count (0 of 5) decides the registered verdict. Dwell: pooled
increment of the dwell pole difference in the same runs, for the true-section
marker over the pairs also valid under the pixel lens, as registered for the
ratio of the pixel-lens increment to the marker increment. Last column: pixel lens on top of the text lens
minus the text lens in attention, seed 0 only. Lower block: registered
re-measurement on the generated-page set (350 pairs each on
sports and fashion, 400 on finance), with seed counts for the marker and
without the last column. There $p$ is below $.05$ on finance at 4B although
the increment is negative, and 13 answers that could not be parsed (12 at 32B)
have attention but no dwell, so 9 Dwell values pool 1 to 3 fewer pairs.}
\label{supptab:words-attn}
\begin{tabular}{@{}llcccccccc@{}}
\toprule
 & & \multicolumn{2}{c}{Text lens} & \multicolumn{3}{c}{Pixel lens} & \multicolumn{2}{c}{True-section marker} & Pixel lens on \\
\cmidrule(lr){3-4}\cmidrule(lr){5-7}\cmidrule(lr){8-9}
Model & Interest & Attn.\ (seeds) & Dwell & Attn.\ (seeds) & $p$ & Dwell & Attn. & Dwell & text lens \\
\midrule
\multicolumn{10}{@{}l}{\emph{Five pages of the weight-lens evaluation set}} \\
4B  & Fash. & $+.001$ (3/5) & $+.184$ & $+.013$ (5/5) & $<.001$ & $+.023$ & $+.018$ & $+.043$ & $+.012$ \\
    & Trav. & $+.010$ (4/5) & $+.192$ & $-.037$ (0/5) & $.011$  & $+.070$ & $-.094$ & $+.042$ & $-.051$ \\
9B  & Fash. & $+.001$ (3/5) & $+.076$ & $-.005$ (1/5) & $.248$  & $+.079$ & $-.071$ & $+.048$ & $-.013$ \\
    & Trav. & $+.010$ (5/5) & $+.160$ & $+.021$ (5/5) & $<.001$ & $+.098$ & $+.034$ & $+.112$ & $+.022$ \\
32B & Fash. & $+.005$ (5/5) & $+.070$ & $+.025$ (5/5) & $<.001$ & $-.050$ & $+.012$ & $+.029$ & $+.020$ \\
    & Trav. & $+.005$ (5/5) & $+.172$ & $-.022$ (0/5) & $.024$  & $+.046$ & $+.064$ & $+.118$ & $-.020$ \\
\midrule
\multicolumn{10}{@{}l}{\emph{Generated-page set (registered re-measurement)}}
\\
4B  & Sports & $-.002$ (2/5) & $+.261$ & $+.042$ (5/5) & $<.001$ & $+.007$ & $+.040$ (5/5) & $+.085$ & --- \\
    & Fash.  & $+.001$ (2/5) & $+.151$ & $+.011$ (5/5) & $<.001$ & $+.032$ & $-.035$ (0/5) & $+.094$ & --- \\
    & Fin.   & $-.002$ (2/5) & $+.180$ & $-.014$ (0/5) & $<.001$ & $+.067$ & $-.0004$ (2/5) & $+.086$ & --- \\
9B  & Sports & $+.010$ (5/5) & $+.164$ & $+.007$ (5/5) & $<.001$ & $+.033$ & $+.003$ (4/5) & $+.082$ & --- \\
    & Fash.  & $+.008$ (5/5) & $+.119$ & $+.009$ (5/5) & $<.001$ & $+.039$ & $+.015$ (5/5) & $+.092$ & --- \\
    & Fin.   & $+.006$ (5/5) & $+.132$ & $-.005$ (0/5) & $.686$ & $+.058$ & $-.009$ (0/5) & $+.097$ & --- \\
32B & Sports & $+.003$ (3/5) & $+.167$ & $+.048$ (5/5) & $.003$ & $+.040$ & $+.120$ (5/5) & $+.060$ & --- \\
    & Fash.  & $+.004$ (4/5) & $+.118$ & $-.051$ (0/5) & $.111$ & $+.029$ & $-.075$ (0/5) & $+.091$ & --- \\
    & Fin.   & $+.011$ (5/5) & $+.134$ & $-.005$ (0/5) & $.753$ & $-.004$ & $-.004$ (0/5) & $+.051$ & --- \\
\bottomrule
\end{tabular}
\end{table*}

\paragraph{Late-layer mask over five seeds.}
The registered statement that the pixel lens acts through the late-layer
attention to the target across seeds is not supported. At 4B with seed 0 its
mean retention over fashion and travel under the late-layer mask is $1.79$, not
below the registered $.5$, and $.48$ under the mask on every layer, the
positive control. In a registered addition with seeds 1 to 4, the criterion of
a travel retention below $.5$ in at least 4 of 5 seeds holds in 2 (seeds 1 and
4) and fails in 3 (seeds 0, 2 and 3, retention $.62$ to $2.00$).
The positive control itself leaves less than half of the travel increment in
three seeds ($.11$ to $.33$) but not in seeds 0 and 4 ($.69$ and $.67$).

\paragraph{Irrelevant attributes.}
The text lens separates personas on attributes that should not move gaze in
12 of 15 tests, against 1 of 15 for the pixel lens and for the true-section
marker. We flipped five attributes that have no matching region on the pages
(Korean food is partly related, since some pages have a food section). The
excess of an attribute is the L1 distance between the dwell of the two poles
minus the seed
noise of that distance, and the ratio divides it by the same quantity for
sports.

The text lens differs from the base prompt in these 12 tests (10 of 12
without Korean food) by 5 to 12\% of its sports effect. The pixel lens on top
of it also differs in 12 of 15. On the other hand, the pixel lens differs only
on knitting at 9B
($+.036$,
$p=.0033$), the true-section marker only on AI for documents at 32B ($+.028$,
$p=.024$, smallest $p=.095$ in the 10 tests at 4B and 9B), and the
marker-naming control in none of its six tests (smallest $p=.077$).

\paragraph{Registered floor check.}
At 4B and 9B the response of the text lens to the irrelevant attributes is not
shown to exceed either of the two floors defined below, and at 32B it exceeds
floor (a) but is not shown
to exceed the stricter floor (b). The mean is $-.005$ ($p=.66$) for floor (a)
and $-.033$ ($p=.999$) for floor (b) at 4B, $+.006$ ($p=.31$) and $-.008$
($p=.89$) at 9B, and $+.030$ ($p<10^{-4}$) and $-.014$ ($p=.83$) at 32B.

Floor (a) writes the note again for the same profile and pole. Floor (b) swaps
the phrase of the flipped attribute with its neighbor in the profile, so only
the order of the phrases changes. Each floor is the same quantity as the
excess, taken between the registered run and the new run of one pole and
averaged over the two poles.

For each page and persona the measure is the
text-lens excess minus its floor, minus the base-prompt excess minus its
floor. We average it per persona over the five attributes and test the 30
persona means with a one-sided Wilcoxon signed-rank test. The response exceeds
a floor when the mean is above zero and $p<.05$, and the verdict of a model
needs both floors. The registered positive control, sports, exceeds both
floors in all three models. Neither registered check of an order effect in
floor (b) has $p<.05$ in any model, and no attribute exceeds both floors in
any model.

The check has little power. We registered it before collecting its data, but
after we had seen a related analysis, added after data collection and not
reported here, of two earlier runs that wrote their notes anew. That analysis
had already inspected the registered runs, so only the floors are new data.
With the floor of that analysis in place of both floors, the response exceeded
both in 6.3,
0.6 and 22.5\% of 20{,}000 bootstrap resamples of the 30 personas at 4B, 9B
and 32B. We computed this before collecting the data, and the registration
calls it optimistic for floor (b).

In an analysis added after data collection, swapping two phrases moves the
text-lens gaze about as much as flipping the attribute does. The text-lens
floor (b) is $+.068$, $+.040$ and $+.077$ at 4B, 9B and 32B, against a
text-lens excess of $+.047$, $+.037$ and $+.068$ over all units (a page, a
persona and an attribute). The ratio is
$1.46$, $1.06$ and $1.14$ ($[1.12, 2.01]$, $[.83, 1.42]$ and $[.96, 1.32]$).
The text-lens floor (b) exceeds the base-prompt floor (b) by $+.07$, $+.04$ and
$+.08$ ($p<10^{-6}$
over the 30 persona means). Therefore, the swap moves the text-lens gaze far
more than the base-prompt gaze. For each attribute and model the base-prompt
floor
(b) lies between $-.006$ and $+.005$ and the text-lens floor (b) between
$+.031$ and $+.088$.

However, the base prompt is not unmoved. At the level of
single units, reported as description and not used for the verdict, the swap,
which moves the finance phrase to the end of the profile, shifts its
finance-section share by $+.015$ at 32B ($p<10^{-12}$).

The registered count
of 12 of 15 tests becomes 10 of 15 with the Benjamini--Hochberg correction
within each model. The registered test repeated in the new runs has
$p<.05$ in 11 of 15 tests with the rewritten note and in 13 of 15 with the
swapped phrases.

\paragraph{Human-likeness.}
The text lens fails the preregistered human-likeness criterion in all eight
comparisons of \cref{supptab:words-ueyes}. The criterion is that the lower
bound of the 95\% confidence interval of the difference to the UEyes
viewers~\cite{ueyes2023chi} (text lens minus base, paired-image bootstrap with
2000 resamples) is at least $-.02$, on the 108 test images, 27 of which are
web pages. The images carry no markers, so both conditions answer with
coordinates, as in Appendix~I.

Two human viewers of one image reach a
ScanMatch of $.421$ over the 108 images and $.427$ over the 27 web pages. The
base values come from their own run and differ slightly from the Base 4B row of
\cref{supptab:human} ($.134$ against $.139$ on the web pages). The 4B
model returns an empty scanpath for 44 of the 108 images with the base prompt
and for 57 with the text lens (9B: 15 and 22), so its comparison rests on fewer
usable scanpaths than the image count suggests.

\begin{table}[t]
\centering\footnotesize
\setlength{\tabcolsep}{2pt}
\caption{Human-likeness on UEyes. CC is the Pearson correlation between the
fixation density maps of the model and of the viewers. Each entry gives the base and text-lens values
and the difference with its 95\% bootstrap interval. Every lower bound is
below the preregistered floor of $-.02$.}
\label{supptab:words-ueyes}
\begin{tabular}{@{}llcc@{}}
\toprule
Model & Set & CC: base$\to$text lens & ScanMatch: base$\to$text lens \\
\midrule
4B & All 108 & $.190\to.152$ & $.115\to.083$ \\
   &         & $-.038$ [$-.071$, $-.004$] & $-.032$ [$-.043$, $-.022$] \\
   & Web 27  & $.146\to.101$ & $.134\to.091$ \\
   &         & $-.045$ [$-.108$, $+.022$] & $-.042$ [$-.058$, $-.027$] \\
9B & All 108 & $.250\to.231$ & $.197\to.179$ \\
   &         & $-.019$ [$-.042$, $+.003$] & $-.018$ [$-.025$, $-.010$] \\
   & Web 27  & $.254\to.223$ & $.222\to.191$ \\
   &         & $-.031$ [$-.078$, $+.022$] & $-.031$ [$-.046$, $-.018$] \\
\bottomrule
\end{tabular}
\end{table}

\paragraph{Marker-naming control.}
The marker-naming control adds more than the pixel lens on the generated and
live pages, and it keeps its effect under the late-layer mask. We ran it as
registered at 4B and 9B. On the generated-page set it adds $.430$, $.301$ and
$.494$ to the pole
difference at 4B and $.224$, $.135$ and $.207$ at 9B (sports, fashion,
finance), more than the pixel lens in 6 of 6 entries. Over the three interests the median
fixation count at 4B is 30 under the base prompt, 15 under the control and 27
under the pixel lens, and at 9B it is 27, 28 and 27.

On the live cells the
control beats both the base prompt and the random marker in 67 and 65 of the
87 cells at 4B and 9B, against 26 and 21 for the pixel lens. It beats the pixel
lens in 66 and 63 cells and loses in 12 and 9. In the attention runs it
raises the target attention by $+.003$ on fashion (3 of 5 seeds) and $+.021$
on travel (4 of 5 seeds), against a dwell increment of $+.565$ and $+.524$. By
the registered rule it moves the target attention on travel and over both
interests (4 of 5 seeds each), but not on fashion. Under the late-layer mask
(seed 0) it keeps $.94$ of its fashion increment and $1.03$ of its travel
increment (mean $.99$), against $.86$ and $.46$ for the text lens (mean
$.66$), above the registered bound of $.5$.

\section{Two Closed Models on the Generated-Page Set and on the Weight-Lens
Evaluation Set}
\label{supp:closed}

We measure the behavior of two closed models of one vendor,
\texttt{gpt-6.1-terra} and \texttt{gpt-6-luna}, whose weights and sizes are
not public. We chose the first because its vendor differs from that of the
assistant named in the LLM Usage Statement of the main paper and its
reasoning can be switched off. A third registered reason, that it accepts the
sampling settings of the open models, did not hold in effect. The second is
a model of the same vendor at one tenth of the price per token. Each model ran
all conditions of Table~1 on the generated-page set, with and without topic
names (Appendix~C), and the base prompt on the
weight-lens evaluation set of Table~3. Each run was registered with the rule
that this appendix reports every result. \cref{supptab:closed-settings}
gives the settings as run and \cref{supptab:closed-lens} every entry on the
generated-page set.
\looseness=-1

\begin{table*}[t]
\centering\footnotesize
\setlength{\tabcolsep}{3pt}
\caption{Settings of the two closed models as run. ``In effect'' follows
from the route check. Costs are estimated from the token counts at the
prices given and, on the generated-page set, include the pilot and the
marker-reading check but not the route check, whose calls are not logged.
The invalid marker numbers and the calls sent again were counted after data
collection.}
\label{supptab:closed-settings}
\begin{tabular}{@{}>{\raggedright\arraybackslash}p{0.17\linewidth}>{\raggedright\arraybackslash}p{0.395\linewidth}>{\raggedright\arraybackslash}p{0.395\linewidth}@{}}
\toprule
Setting & \texttt{gpt-6.1-terra} & \texttt{gpt-6-luna} \\
\midrule
Model name returned & The requested name in all 31{,}813 successful calls & The requested name in all 31{,}866 successful calls \\
Access & \multicolumn{2}{>{\raggedright\arraybackslash}p{0.80\linewidth}@{}}{Institutional API gateway that forwards requests to a third-party relay, for both models} \\
Date & 11 November 2026. Design calls on the generated-page set within 129
minutes & 11 November 2026. Design calls on the generated-page set within 179
minutes, including 65 minutes with no call around a restart \\
Price per million tokens & \$1 input and \$5 output, the rates of the gateway's cost report & \$0.10 input and \$0.50 output, published prices not checked against the gateway \\
Reasoning & Requested off. Reasoning tokens in 15 of 28{,}800 gaze calls, 0 of 2{,}760 note calls and 1 of 13 calls of the marker-reading check, 5{,}460 tokens in all & Requested off. No reasoning tokens in any call \\
Route check & One input-token count (33) and 24 different answers in 24 identical requests. Distinct values in 24 random-integer requests: 7 at temperature 0, 10 at 2.0, 8 with no setting and 1 at top-$p$ $0.01$ & One input-token count (32) and 24 different answers in 24 identical requests. Distinct values in 24 random-integer requests: 6 at temperature 0, 3 at 2.0, 5 with no setting and 1 at top-$p$ $0.01$ \\
Scanpath sampling & \multicolumn{2}{>{\raggedright\arraybackslash}p{0.80\linewidth}@{}}{Requested temperature $0.7$ and top-$p$ $0.95$. In effect the server's default temperature and top-$p$ $0.95$, for both models} \\
Note of the text lens & \multicolumn{2}{>{\raggedright\arraybackslash}p{0.80\linewidth}@{}}{Requested temperature 0. In effect one sample at the server's default temperature with no top-$p$, for both models} \\
Seeds & \multicolumn{2}{>{\raggedright\arraybackslash}p{0.80\linewidth}@{}}{None. The two repeats are separate requests} \\
Image & \multicolumn{2}{>{\raggedright\arraybackslash}p{0.80\linewidth}@{}}{$1280\times800$ PNG, detail ``high'', preprocessed by the server} \\
Output caps & \multicolumn{2}{>{\raggedright\arraybackslash}p{0.80\linewidth}@{}}{1000 tokens (scanpath), 150 tokens cut at 600 characters (note)} \\
Calls on the generated-page set & 28{,}800 gaze and 2{,}760 note calls, plus
240 pilot calls and 13 calls of the marker-reading check & 28{,}800 gaze
trials, of which 52 were refused, and 2{,}760 note calls, plus 240 pilot calls
and 6 calls of the marker-reading check. 111 answered gaze calls of stopped or
cut batches were sent again \\
Refusals and parsing & No refusal. All 28{,}800 gaze answers parse & The refusals that stopped four of the six jobs were not logged. After the fix the vendor's policy classifier refused 52 gaze calls. No note call was refused. 28{,}724 of 28{,}800 gaze rows parse \\
Invalid marker numbers & 3 of 580{,}167 & 132 of 626{,}075 \\
Retries & 39 (32 after an HTTP 502 error, 6 after a time-out, 1 after an HTTP 500 error), each successful on the second attempt & 61 (60 after an HTTP 502 error, 1 after an empty answer), each successful on the second attempt \\
Tokens and cost & $49.2$M input and $6.6$M output tokens, \$82.13 & $49.3$M input and $7.0$M output tokens, \$8.44 \\
Weight-lens evaluation set & 840 gaze calls under the requested name, 1 retry,
no refusal, no reasoning tokens, \$2.25, plus 48 pre-check calls for \$0.12 &
840 gaze calls under the requested name, no retry, no refusal, no reasoning
tokens, \$0.22, plus 48 pre-check calls for \$0.01 \\
\bottomrule
\end{tabular}
\end{table*}

\begin{table*}[t]
\centering\footnotesize
\setlength{\tabcolsep}{4pt}
\caption{Entries of Table~1 of the main paper for the two closed models on
the pages with and without topic names, with two further registered columns.
Base prompt is the pole difference with its 95\% bootstrap interval over the
pairs. The other columns are increments over it, except Pixel$-$rand.\
(pixel lens minus random marker) and Text$-$pixel (text lens minus pixel
lens). True sect.\ is the true-section marker and rec.\ the recovery. A pair
is a page, a persona and a repeat (8 pages, 30 personas and 2 repeats per
interest), and all conditions of a page set come from one run. Pairs counts
the pairs of the base prompt. For \texttt{gpt-6-luna}, refused requests and
answers that do not parse remove pairs, fewest under the text lens, and the
other columns keep 470 to 479 pairs on all pages and 410 to 419 on the
generated pages. ``Generated only'' leaves out the live Naver page, which is
the same in both sets. A star marks one-sided paired Wilcoxon $p<.05$.
``Without $-$ with'' is the registered paired difference of the base-prompt
pole difference between the sets on the generated pages, and a dagger marks
two-sided $p<.05$.}
\label{supptab:closed-lens}
\begin{tabular}{@{}llrlccccc@{}}
\toprule
Interest & Pages & Pairs & Base prompt [95\% CI] & True sect. & Pixel lens (rec.) & Pixel$-$rand. & Text lens & Text$-$pixel \\
\midrule
\multicolumn{9}{@{}l}{\texttt{gpt-6.1-terra}} \\
Sports  & With topic names     & 480 & $+.461^{\star}$ [.442, .479] & $+.019^{\star}$ & $+.015^{\star}$ (.79) & $+.022^{\star}$ & $+.044^{\star}$ & $+.029^{\star}$ \\
        & Without topic names  & 480 & $+.370^{\star}$ [.355, .385] & $+.026^{\star}$ & $+.016^{\star}$ (.61) & $+.014^{\star}$ & $+.071^{\star}$ & $+.055^{\star}$ \\
        & Without $-$ with     & 480 & $-.091^{\dagger}$ & & & & & \\
\addlinespace[2pt]
Fashion & With topic names     & 480 & $+.292^{\star}$ [.272, .312] & $+.015^{\star}$ & $+.009^{\star}$ (.63) & $+.014^{\star}$ & $+.049^{\star}$ & $+.040^{\star}$ \\
        & \quad generated only & 420 & $+.297^{\star}$ [.275, .319] & $+.013^{\star}$ & $+.009^{\star}$ (.68) & $+.011^{\star}$ & $+.040^{\star}$ & $+.032^{\star}$ \\
        & Without topic names  & 480 & $+.234^{\star}$ [.215, .254] & $+.023^{\star}$ & $+.016^{\star}$ (.72) & $+.019^{\star}$ & $+.050^{\star}$ & $+.034^{\star}$ \\
        & \quad generated only & 420 & $+.233^{\star}$ [.213, .254] & $+.017^{\star}$ & $+.016^{\star}$ (.95) & $+.019^{\star}$ & $+.040^{\star}$ & $+.024^{\star}$ \\
        & Without $-$ with     & 420 & $-.064^{\dagger}$ & & & & & \\
\addlinespace[2pt]
Finance & With topic names     & 480 & $+.416^{\star}$ [.400, .432] & $+.035^{\star}$ & $+.021^{\star}$ (.60) & $+.021^{\star}$ & $+.083^{\star}$ & $+.062^{\star}$ \\
        & \quad generated only & 420 & $+.436^{\star}$ [.419, .453] & $+.036^{\star}$ & $+.020^{\star}$ (.55) & $+.020^{\star}$ & $+.073^{\star}$ & $+.053^{\star}$ \\
        & Without topic names  & 480 & $+.328^{\star}$ [.314, .341] & $+.032^{\star}$ & $+.020^{\star}$ (.64) & $+.022^{\star}$ & $+.088^{\star}$ & $+.068^{\star}$ \\
        & \quad generated only & 420 & $+.331^{\star}$ [.316, .346] & $+.036^{\star}$ & $+.024^{\star}$ (.66) & $+.022^{\star}$ & $+.082^{\star}$ & $+.058^{\star}$ \\
        & Without $-$ with     & 420 & $-.104^{\dagger}$ & & & & & \\
\midrule
\multicolumn{9}{@{}l}{\texttt{gpt-6-luna}} \\
Sports  & With topic names     & 478 & $+.178^{\star}$ [.161, .195] & $+.062^{\star}$ & $+.040^{\star}$ (.65) & $+.079^{\star}$ & $+.272^{\star}$ & $+.232^{\star}$ \\
        & Without topic names  & 475 & $+.066^{\star}$ [.056, .076] & $+.056^{\star}$ & $+.044^{\star}$ (.79) & $+.060^{\star}$ & $+.260^{\star}$ & $+.214^{\star}$ \\
        & Without $-$ with     & 473 & $-.113^{\dagger}$ & & & & & \\
\addlinespace[2pt]
Fashion & With topic names     & 475 & $+.099^{\star}$ [.086, .113] & $+.046^{\star}$ & $+.013^{\star}$ (.29) & $+.026^{\star}$ & $+.174^{\star}$ & $+.162^{\star}$ \\
        & \quad generated only & 415 & $+.104^{\star}$ [.091, .119] & $+.037^{\star}$ & $+.022^{\star}$ (.60) & $+.029^{\star}$ & $+.150^{\star}$ & $+.129^{\star}$ \\
        & Without topic names  & 477 & $+.050^{\star}$ [.041, .058] & $+.031^{\star}$ & $+.001$\hphantom{$^{\star}$} (.04) & $+.017^{\star}$ & $+.154^{\star}$ & $+.151^{\star}$ \\
        & \quad generated only & 417 & $+.046^{\star}$ [.039, .053] & $+.012^{\star}$ & $+.009^{\star}$ (.69) & $+.019^{\star}$ & $+.125^{\star}$ & $+.114^{\star}$ \\
        & Without $-$ with     & 412 & $-.059^{\dagger}$ & & & & & \\
\addlinespace[2pt]
Finance & With topic names     & 479 & $+.152^{\star}$ [.137, .167] & $+.041^{\star}$ & $+.022^{\star}$ (.54) & $+.052^{\star}$ & $+.212^{\star}$ & $+.191^{\star}$ \\
        & \quad generated only & 419 & $+.159^{\star}$ [.142, .176] & $+.038^{\star}$ & $+.017^{\star}$ (.45) & $+.053^{\star}$ & $+.214^{\star}$ & $+.199^{\star}$ \\
        & Without topic names  & 479 & $+.088^{\star}$ [.078, .099] & $+.044^{\star}$ & $+.027^{\star}$ (.62) & $+.039^{\star}$ & $+.135^{\star}$ & $+.109^{\star}$ \\
        & \quad generated only & 419 & $+.084^{\star}$ [.072, .095] & $+.046^{\star}$ & $+.028^{\star}$ (.60) & $+.040^{\star}$ & $+.131^{\star}$ & $+.104^{\star}$ \\
        & Without $-$ with     & 418 & $-.076^{\dagger}$ & & & & & \\
\bottomrule
\end{tabular}
\end{table*}

\paragraph{Access and settings.}
Both models were reached through an institutional API gateway that forwards
requests to a third-party relay, so we cannot verify independently
which model served the requests, although every successful call returned the
requested name. The names carry no dated version and the API takes no seed, so
a rerun cannot be identical. A route check with 24 identical requests and 24
requests per
sampling setting found no sign of stored answers. Identical requests gave
different answers, and in counts added after data collection none of the
14{,}400 and 14{,}324 repeat pairs with the same page, persona, pole and
condition have identical scanpaths. The check was registered for the second
model only. For the first it ran when the pages with topic names were complete
and about two thirds of those without had been collected. Before the
registration
for the second, an unrecorded check without stimuli had shown that its name
accepts the settings, and a first attempt of the route check with reasoning
left on had returned errors.

For both models the temperature is accepted without an error but not
applied, while top-$p$ is applied, so the settings in effect
(\cref{supptab:closed-settings}) differ from those requested, and we cannot
tell whether the relay or the model drops the temperature. The check before
the first registration had tested only that the settings are accepted. Since
the temperature is not applied, each note is one sample, and both repeats reuse
it. The first model finished the pages without topic names with the same
settings, the notes were not rerun, and within a
run all five conditions were sampled alike, so each registered comparison is
made under one setting.

The reasoning tokens of the first model are below the registered limit of
1\% of the calls, and these answers are kept.
\looseness=-1

\paragraph{Input checks of the first model.}
\texttt{gpt-6.1-terra} failed the first version of the registered
marker-reading check and passed an amended version, which departs from the
registered rule. The registered criterion was $.80$ correct on 48 text
snippets from six pages, and a model that fails it was not to be run. The
first version gave 30 of 48 ($.625$). In a count added after data collection,
16 of the 18 wrong answers name a tag that overlaps the element of the
snippet, so these questions had no single correct answer. We did not drop the
model. Before the pilot and before any design call (a gaze or note call of the
registered runs), we amended the check to
skip a snippet when its tag overlaps another element or another tag overlaps
it. Of the 25 snippets left, the model answered 24 ($.96$) correctly. A pilot
outside the design then passed its registered criteria, and its gaze results
were not used.

By an error, the main runs of both models reused marker screenshots of the
pilot from the image cache for 10 personas on 2 pages under sports. In a
check added after data collection, this concerns 108 of the 28{,}800 gaze
trials of each model. Without the 18 affected pairs no sports entry with
topic names changes by more than $.004$, and every star remains. Both models
have 120 fewer note calls than registered, because the live Naver page is the
same in both page sets and the set without topic names reused its notes, as in
the open-model
runs.
\looseness=-1

\paragraph{Run of the second model.}
\texttt{gpt-6-luna} passed the route check, the amended marker-reading check
(24 of 25) and the pilot, but in its main run the registered rule for
refused requests was not applied to some of them. The runner did not recognize
refusals by the vendor's policy classifier and stopped four of its six jobs,
and a restart of our computer cut the other two. The registered rule treats a
refused request as a missing trial and removes both poles of its pair.

We
fixed the implementation, not the rule or the settings, and sent the unwritten
trials of the stopped or cut batches again, including the requests whose
refusal had stopped the jobs. In counts added after data collection, 111 of
the trials sent again had already been answered once, and at least 2 refused
requests were answered when sent again. Removing the four batches in which a
refusal had stopped a job changes no star and no reported entry by more than
$.005$.

After the fix the classifier refused 52 gaze calls. With 24 answers
that the parser could not read, 76 of the 28{,}800 gaze rows do not parse,
and the result rows do not tell the two kinds apart. As registered during the
run, by condition there are 19 under the true-section marker, 18 under the
pixel lens, 18 under the base prompt, 17 under the random marker and 4 under
the text lens, and by pole 43 Passionate and 33 Averse rows.
\looseness=-1

\paragraph{Results on the generated-page set.}
Each of the three registered direction predictions holds in 6 of 6 entries
(two page sets times three interests) for each model. The base-prompt pole
difference is positive, the pixel lens is above the random marker, and the
text lens is above the pixel lens on the same pairs, all with one-sided
$p<.05$.

Of the entries registered without a prediction, the true-section
marker and the text lens are significant in 6 of 6 for both models. The pixel
lens is significant in 6 of 6 for \texttt{gpt-6.1-terra} and 5 of 6 for
\texttt{gpt-6-luna}, where fashion without topic names adds $+.001$
($p=.19$). In a count added after data collection, the random marker alone
stays below the base prompt in 6 of 6 entries for \texttt{gpt-6-luna}, by
$.012$ to $.038$ (two-sided $p<.05$ in 4 of 6).
In the registered paired comparison between the page sets, the base-prompt
pole difference on the generated pages is lower without topic names on every
interest for both models (two-sided $p<10^{-50}$ for \texttt{gpt-6.1-terra}
and $p<10^{-14}$ for \texttt{gpt-6-luna}).
\looseness=-1

\paragraph{Base prompt on the weight-lens evaluation set.}
The registered prediction for the closed-model rows of Table~3 of the main
paper, a positive Held-out $\Delta$ on both interests for both models in a
one-sided paired Wilcoxon test at $p<.05$ without correction, holds in 4 of
4 entries (largest $p=1.6\times10^{-11}$). Conc.\ had no prediction.

The
rows come from a second registered run of the base prompt on the weight-lens
evaluation set of Sec.~4.4 of the main paper, in the configuration of the
open-model base rows. Their placement in Table~3 was decided before the
results. At this registration the results on the generated-page set were
known, those of \texttt{gpt-6-luna} for about four minutes, but the runner, the
analysis script and the draft of the registration were written
before them. A registered pre-check of 48 calls per model, with personas
outside the design, passed for both. The
95\%
bootstrap intervals over the pairs (40 on fashion, 80 on travel) are
$[.523, .586]$ and $[.359, .478]$ for \texttt{gpt-6.1-terra} and
$[.235, .296]$ and $[.144, .214]$ for \texttt{gpt-6-luna}.

On page rp-004 the travel section lies below the viewport and carries no tag.
Therefore, its 20 travel pairs are zero for both poles in every model, open or
closed, and the travel value includes them, as in Table~3. Two differences
from the open models remain. First, we could not check whether their machines
drew the tags with the same font and image library. Second, their scripts
resize the screenshot before the image processor ($1280\times800$ becomes
$1288\times812$), while the closed models received the $1280\times800$ PNG.

\end{document}